\documentclass[letterpaper]{article} 
\usepackage{aaai25}  
\usepackage{times}  
\usepackage{helvet}  
\usepackage{courier}  
\usepackage[hyphens]{url}  
\usepackage{graphicx} 
\usepackage{natbib}  
\usepackage{caption} 
\usepackage{algorithm}
\usepackage{algorithmic}
\usepackage[table]{xcolor}
\usepackage{newfloat}
\usepackage{listings}
\DeclareCaptionStyle{ruled}{labelfont=normalfont,labelsep=colon,strut=off} 
\floatstyle{ruled}
\newfloat{listing}{tb}{lst}{}
\floatname{listing}{Listing}
\usepackage{listings}
\usepackage{algorithm}
\usepackage{algorithmic}
\usepackage{multirow}
\usepackage{csquotes}
\usepackage{subcaption}
\usepackage{multirow}
\usepackage{booktabs}
\usepackage{graphicx}
\usepackage{rotating}
\usepackage{colortbl}
\usepackage{cleveref}
\newcommand{\answerYes}[1]{\textcolor{black}{#1}}

\title{Hatred Stems from Ignorance! Distillation of the Persuasion Modes in Countering Conversational Hate Speech }
\author{
    Ghadi Alyahya, Abeer Aldayel \\
}
\affiliations{
    King Saud University, 
College of Computer and Information Sciences\\

    ghadii55@gmail.com, aabeer@ksu.edu.sa 
   
}

\usepackage[switch]{lineno}  

\begin{document}
\maketitle
\footnotetext{The title “Hatred Stems from Ignorance” reflects the study’s theoretical grounding in Aristotle’s \textit{Rhetoric}, as interpreted by Cope, E. M., and Sandys, J. E. (2010). This foundation aligns with the insights of Ibn Rushd (Averroes), a renowned philosopher known for reintroducing Aristotle’s texts. The phrase is inspired by one of Ibn Rushd’s reflections on the roots of hatred and ignorance.}
\begin{abstract}
Examining the factors that the counterspeech uses are at the core of understanding the optimal methods for confronting hate speech online. Various studies have assessed the emotional base factors used in counter speech, such as emotional empathy, offensiveness, and hostility. To better understand the counterspeech used in conversations, this study distills persuasion modes into reason, emotion, and credibility and evaluates their use in two types of conversation interactions: closed (multi-turn) and open (single-turn) concerning racism, sexism, and religious bigotry. The evaluation covers the distinct behaviors seen with human-sourced as opposed to machine-generated counterspeech. It also assesses the interplay between the stance taken and the mode of persuasion seen in the counterspeech. 

Notably, we observe nuanced differences in the counterspeech persuasion modes used in open and closed interactions—especially in terms of the topic, with a general tendency to use reason as a persuasion mode to express the counterpoint to hate comments. The machine-generated counterspeech tends to exhibit an emotional persuasion mode, while human counters lean toward reason. Furthermore, our study shows that reason tends to obtain more supportive replies than other persuasion modes. The findings highlight the potential for incorporating persuasion modes into studies about countering hate speech, as they can serve as an optimal means of explainability and pave the way for the further adoption of the reply's stance and the role it plays in assessing what comprises the optimal counterspeech.

\end{abstract}

\section{Introduction}
\label{sec:intro}
Counterspeech is a strategy that is commonly adopted to combat online hate speech~\citep{Parker2023-cy}. The dependency on counterspeech as a main component to confront hate has been further intensified by the prevalence of online conversation-based platforms~\citep{Caldevilla-Dominguez2023-hl}, either multi-turn conversation platforms (chat systems that provide closed interactions between two users, closed (multi-turn)) or single-turn conversation platforms (e.g., social media platforms, where many users can be involved in a conversation in an open interaction setting, open (single-turn)). Multiple studies have begun directing attention to adopting counterspeech strategies to halt the spread of hate and alleviate the effects of blocking or deleting the content on freedom of speech~\citep{Gagliardone2015-qf,Myers_West2018-up,Garland2020-gv}. 

However, counterspeech still faces multifaceted challenges due to its dependency on content dynamics and the toll of the interaction type – i.e., closed or open~\citep{Catherine2022-os}. These challenges intensify with the recent adoption of machine-generated counterspeech to limit human moderation costs~\citep{Ashida2022-dd}. Previous studies have addressed the dynamics of counterspeech content by analyzing words' emotional connotations and sentiment polarity~\citep{Brassard-Gourdeau2019-iu,Garland2020-gv} or empathy~\citep{Hangartner2021-up,Lahnala2022-oa}. Similarly, current studies have begun investigating content dynamics in the context of machine-generated and human-sourced narratives. For instance,~\citet{Bonaldi2022-yj} and~\citet{Sen2023-le} compare human-sourced and machine-generated counterspeech. Still, current attempts to distill the non-emotional aspects of counterspeech are minimal. They typically analyze one target type in the argument strategy, such as stereotype-targeted strategies~\citep{Mun2023-gv}, with a limited focus on one interaction type and ignore replies to the counterspeech~\citep{Mathew2019-or}.

The persuasion mode is a communication method involving multiple parties who exchange various argu- ments to persuade the other participants of the merits of a different perspective. This is done in an attempt to strengthen their convictions~\citep{Barden2012-sg,Dutta2020-on}. \textcolor{black}{To define the set of persuasion modes for the analysis, we used the conceptual aspects of a persuasive argument based on the rhetorical appeals of persuasion theory, which categorizes argumentative building blocks into logos (reason), pathos (emotion), and ethos (credibility)~\citep{Lukin2017-bu,Cope2010-gf}. Further, counterspeech is a direct response to hate speech and a form of intervention that confronts the hateful content and refutes the argument to prove the counterpoint~\citep{Schmuck2024-mc,Gagliardone2015-qf}. We hypothesize that persuasion modes may appear in counterspeech and hate-speech contexts, as people confronting hateful arguments tend to use persuasive appeals to convince others with their counterpoints.}

In this study, we distilled persuasion modes used in counterspeech, focusing on two main dimensions: interaction type and context-based behaviors (human-based or machine-generated contexts). Particularly, this study analyzed the persuasion modes used in counterspeech as well as their interplay with hate speech in conversations.

Our research questions (RQs) are as follows:
\begin{itemize}
\item RQ1: How effective are the persuasion modes in identifying counterspeech across domains (or topics) and interaction types?
\item RQ2: Does the persuasion mode used in counterspeech vary by interaction type and the content of the machine-generated and human-written conversation? 
\item RQ3: To what extent do counterspeech persuasion modes drive the follow-up replies in conversations?

\end{itemize}
We perform experiments using a subset of two multi-turn, conversation-based counterspeech datasets~\citep[DialogConan, ][]{Bonaldi2022-yj}, which contains approximately 8,000 multi-turn (closed interaction) conversations between two users and 1,500 single-turn~\citep[ContextCounter, ][]{Albanyan2023-fc} \textit{X} posts demonstrating open interactions among many users \footnote{Throughout the paper, we use the terminology closed (multi-turn) to indicate the closed conversations' interactions that contain multi-turns between two users, such as chat online systems. We use open (single-turn) to indicate open conversations' interactions that contain a single turn with many user interactions, such as social media platforms.  }
. Persuasion mode annotations were added to each turn in the conversation for hate speech and counterspeech. We extended the experiment by detailing the type of persuasion mode used in counterspeech as presented by humans compared to the generative model. We also investigated the context of the replies to the counterspeech as another discourse level that may serve as a proxy for understanding the persuasion mode's dimensionality and inferring the counternarratives confronting the hate in these conversations.  

This paper's specific contributions include the following:
\begin{itemize}
\item We provide a coarse-grain examination of the persuasion modes used in hate speech and counterspeech based on emotion, reason, and credibility.
\item We empirically demonstrate the efficacy of incorporating the persuasion mode while detecting hate speech and counterspeech for closed and open conversational interactions.
\item We assess the persuasion modes in machine-generated counterspeech. Considering the tendency of the current studies to adopt machine-generated counterspeech to mitigate hate in conversations~\citep{Ashida2022-dd}, we further analyze the persuasion modes by comparing human-written and machine-generated counterspeech to highlight effective counterspeech strategies for various domains.
\item We leverage the distinct features of single-turn conversations created by allowing multiple users to interact with counterspeech in order to extend the analyses and evaluate the extent to which the replies support the counterspeech. We do so by disentangling four types of follow-up replies: supporting the counterspeech, opposing the counterspeech, supporting the counterspeech with additional context, and opposing the counterspeech with hateful content.
\end{itemize}
\textcolor{black}{This work provides general insights into how the three persuasion modes (emotion, reason, and credibility) used in counterspeech mitigate hate. This aim is to increase the focus on persuasion modes in the computational modeling of conversations and its potential to disentangle the counterspeech's effectiveness in mitigating hateful conversations.}
\section{Previous Work}
\label{sec:previous_work}
Counterspeech to hateful conversations is an argumentative appeal that attempts to challenge and contrast the hateful point~\citep{Gagliardone2015-qf}. There have been attempts to study 
counterspeech by looking at the aspects of the content or conversation behaviors (human as opposed to machine-generated)~\citep{Gupta2023-oi,Sen2023-le}. In this section, we detail related works by the type of the content aspect that they explored. Essentially, aspect refers to the core component of the counterspeech that the study investigated, which can be further categorized as emotion-based and non-emotion-based. Following this, we discuss recent persuasion identification work and list the common techniques used in the field.
\subsection{Counterspeech Aspects}
Previous studies have thoroughly investigated \textbf{the emotion-based aspects of counterspeech}. This is partly due to the ubiquity of the connotation-based tool that facilitates word connotation analysis. The study by~\citep{Hangartner2021-up} shows that using empathy as a shaming technique in counterspeech causes hateful commenters to delete their posts. This work analyzed approximately 1,350 X (formerly Twitter) users and shows that empathy-based counterspeech increases the retrospective deletion of hateful comments. Similarly, the work by~\citep{Lahnala2022-oa} thoroughly examined the use of empathy to mitigate toxicity and shows a distinction between emotional and cognitive empathy. This finding shows that empathy's cognitive components significantly outperform its emotional components in toxicity mitigation. Further, sentiment is widely employed as a proxy for identifying hateful comments. For instance, the work by~\citep{Brassard-Gourdeau2019-iu} used sentiment detection to recognize toxic and subversive online comments. Another work by~\citep{Yu2022-zb} studied the sentiments in content and sarcasm as a figurative language using six pairs of parent and target Reddit comments (original Reddit posts and target comments). 
The Study then evaluated the text polarity's effectiveness and figurative language in identifying hate speech and counterspeech. It found that counterspeech contains fewer negative and derogatory words than hateful comments.

Conversely, there have also been attempts to study \textbf{the non-emotion-based aspects of counterspeech}.
In this setting, the content's behavior and socio-linguistic features are the main aspects of counterspeech analysis. For instance, the work by~\citep{Mun2023-gv} analyzed strategies used in the counterspeech to condemn the underlying inaccurate stereotypes used in hateful comments. Their work analyzed the following six psychological strategies: alternate groups, external factors, alternate qualities, counterexamples, broadening, and general denouncing. They used posts 
from the CONAN dataset and approximately 6,000 Reddit posts to evaluate human preferences for these strategies in human-written and machine-generated conversations. Their findings show that human-written countering strategies are specific to the implied stereotype, whereas machine-generated counterspeech uses general strategies. A study by~\citep{Mathew2019-or} defines counterspeech types based on the strategy employed by users: presenting fact, warning of offline or online consequences, using humor, and showing hostility. They analyzed 13,000 YouTube comments and found that "hostile language" is the major strategy in most instances of counterspeech.
\subsection{Content Dynamics and Interaction Types}
Recent works have explored generated counterspeech using the instruction-based large language models (LLMs)
to mitigate hate speech. For instance, the work by~\citep{Sen2023-le} introduces a semi-automated generation mechanism using GPT-2, GPT-3.5, and FLAN-T5 to produce counterspeech to combat hateful comments, specifically targeting sexism and general hate. A work by~\citep{Agarwal2023-lo} uses Llama 2 and GPT-3.5 to rephrase hateful comments and mitigate the messages' offensiveness. However, researchers have also begun evaluating the effectiveness of off-the-shelf, black-box hate speech detection models. For instance,~\citep{Pozzobon2023-dp}'s work describes the difficulties encountered while utilizing black-box APIs for the evaluation of toxicity in research. In fact, black-box APIs are software tools that provide access to pre-trained models without revealing their internal operations. APIs facilitate advanced model use; however, they introduce significant challenges, including the lack of transparency in the models' decision-making processes. Validating a model's result can be difficult without a thorough understanding of how the model made its predictions.

\subsection{Persuasion Techniques in Conversations}
Persuasion is a communication technique that attempts to influence or convince others of a specific action or conclusion on an issue~\citep{Dutta2020-on}. It has been incorporated into different conversation-based processing scenarios and has been used in counternarrative speech to influence speakers~\citep{Wright2017-df}. However, it is primarily used in fake news detection to identify conversation amplifications~\citep{Orru2022-mf}.  

The work by ~\citep{Pauli2022-ao} models persuasion by using rhetorical appeals to find undesired persuasion in the spread of misinformation. Aristotle defines rhetoric as the study of persuading or influencing others through speech or writing that utilizes three types of appeals: logos, ethos, and pathos~\citep{Rapp2023-pl,Cope2010-gf}. To elaborate, ethos is the process of persuading listeners using the speaker's credibility, while pathos persuades using listeners' emotions, and logos persuades them using the argument's validity. ~\citep{Pauli2022-ao}'s findings demonstrate that the misuse of rhetorical appeals is prevalent in misinformation.
\textcolor{black}{More recently, a couple of studies examined persuasion in closed (multi-turn) conversations (more specifically, conversation generated by language models)~\citep{Salvi2024-jd,Durmus2024-ho}. In these studies, persuasion was viewed as a fixed factor representing the degree of conviction in the conversation, indicating whether individuals were convinced or not.
These studies' outcomes confirm the key role that persuasive language plays in human conversations. Beyond the rigid examination of persuasion as an outcome of a convincing conversation, we provide a deeper examination of the type of persuasion based on the rhetorical appeal, using three primary modes of argument as defined by Aristotle~\citep{Lukin2017-bu,Cope2010-gf}. Working with the little existing empirical evidence about modeling persuasive language in hateful conversations, this study assessed the incorporation of persuasion as a factor of hate speech detection and examined its interplay with counternarratives.}

\begin{table} [!h]
    \centering
  
    \begin{tabular}{c|cc|c}  
 \multicolumn{4}{c}{\textbf{Closed (multi-turn)}  }\\
 Topic& HS
& CS
&Total\\\hline \hline 
         Racism&  1,480&  1,477& 2,957\\ 
         Sexism&  1,390&  1,391& 2,781\\ 
         Religious bigotry &  1,437&  1,437& 2,874\\ 
 \multicolumn{3}{c}{}&8,612\\ \hline 
         \multicolumn{4}{c}{\textbf{Open (single-turn)}  }\\
 Topic& HS
& CS
&Total\\ \hline \hline 
         Racism&  211&  297& 508\\ 
         Sexism&  221&  328& 549\\ 
         Religious bigotry&  125&  154& 279\\
         Other&  108&  143& 251\\ \hline
 \multicolumn{3}{c}{}&1,587\\ \hline
    \end{tabular}
    \caption{Distribution of hate speech (HS) and counterspeech (CS) for the Open (single-turn) and Closed (multi-turn) datasets.}
    \label{tab:datasets_hs_cs_dis}
\end{table}
\section{Experimental Setup}


\subsection{Counterspeech Baseline Datasets}
\label{sec:dataset_Experemnt}

To conduct our experiment, we selected two recent and well-known counterspeech datasets. For the closed dataset, we used ~\citep[DialogConan,][]{Bonaldi2022-yj}, which contains approximately 3,000 unique dialogs between two users and nearly 16,000 instances of turn-taking in conversations covering different topics spanning the following three domains: racism, sexism, and religious bigotry. For open interactions, we selected the recently released dataset ~\citep[ContextCounter, ][]{Albanyan2023-fc}, which contains 2,000 \textit{X} posts comprising interactions among many users. These datasets provide a solid dialogical data baseline and support our experiment's aim to investigate the interaction type and context of replies. Further, ContextCounter is a context-based dataset taken from \textit{X}, a social media platform containing hate speech, counterspeech, and replies to counterspeech. This dataset further focuses on identifying replies' stances toward and the type of context they add to counterspeech in \textit{X posts}. To manage the gold annotation (labels by trained human annotators) costs, we selected half of the dialogs from DialogConan in each topic. 

Adhering to the conversational structure observed in multi-turn conversations, whose average length was seven or more turns, the final gold annotation count for single-turn conversations reached 1,500. The multi-turn conversation dataset has several topic categories. We merged the fine-grained topic labels to unify the hate speech labeling process in the closed conversation dataset. Meanwhile, the open (single-turn) conversation dataset lacked topic labels. Therefore, human annotators labeled the data by conversation topic and persuasion modes (for more details on the data preprocessing, Appendix A). Table ~\ref{tab:datasets_hs_cs_dis} illustrates the distribution of the two datasets that we used for our experiments concerning hate speech and counterspeech, spanning the three topics \footnote{\url{https://github.com/AbeerAldayel/counterspeech-persuasion-modes} }.
\subsubsection{Persuasion Mode Labeling}
\label{sec:persLabeling}
\textcolor{black}{For each conversation set, the annotation has been carried out using Labelbox\footnote{\url{https://labelbox.com/}}. The annotator's team has been recruited using Boost Workforce, which is offered by LabelBox, for more specialized labelers with expertise that matches the annotation use case. The annotators were guided by specific annotation instructions to define persuasion labels for multi-turn and single-turn conversations and to identify the conversation topic of single-turn ones. (Appendix B provides further details on the annotation process.) Each turn of the conversation has been labeled by two annotators following specific tasks of the labeling workflow \footnote{\url{https://docs.labelbox.com/docs/workflows}}. In the first phase, to ensure the annotation's quality, annotators have to first undertake a quality test on a subset of the dataset (Initial labeling task). Then, a third annotator (one of the authors) works on verifying the quality of the annotations and reviewing the labeling results in case of disagreement between the annotators before finalizing the set of labels (review and rework tasks).  
The overall average agreement between the annotators for each topic was around 70\%, demonstrating optimal annotation quality.} 

\textcolor{black}{We used the conceptual aspects of persuasive arguments based on persuasion theory's rhetorical appeals of ethos, logos, and pathos, to define persuasion mode labels ~\citep{Lukin2017-bu,Cope2010-gf}.} This type of labeling provides a method for understanding the high-level modes of persuasion in counterspeech and its interplay with hate speech. Since this study focuses on the persuasion modes used in counterspeech, we used coarse-grained persuasion labels rather than fine-grain strategies. Moreover, research has begun grouping persuasion strategies by their original conceptual definitions. For example, ~\citep{Pauli2022-ao} grouped the fake news detection fallacy types into faulty ethos, logos, and pathos appeals. In this study, we implemented high-level persuasion distillation using coarse-grain persuasion modes to categorize these three persuasion modes as reason (logos), credibility (ethos), and emotion (pathos). Each label is defined in detail as follows: 
\begin{itemize}
\item Reason: This label indicates logos as a mode of persuasion. It appeals to the utilization of logic or reason (e.g., presenting related examples or supporting evidence). 
\item Credibility: This label indicates ethos as a mode of persuasion. It relies on the credibility or authority derived from the speaker’s personal experience.
\item Emotion: This label indicates pathos as a mode of persuasion. It evokes the emotions and feelings of the audience or puts them in a certain mood appealing to their emotions.
\end{itemize}
\begin{table*}[ht!]
\centering
\small
\renewcommand{\arraystretch}{1.2}
\setlength{\tabcolsep}{5pt}

\begin{tabular}{l l l r r r r r r r r r r}
\toprule
\multicolumn{3}{l}{} 
& \multicolumn{3}{c}{\cellcolor{gray!10}Sexism} 
& \multicolumn{3}{c}{\cellcolor{gray!10}Racism} 
& \multicolumn{3}{c}{\cellcolor{gray!10}Religious Bigotry} 
& \cellcolor{gray!10}Macro Avg. \\
\cmidrule(lr){4-6} \cmidrule(lr){7-9} \cmidrule(lr){10-12}
\multicolumn{3}{l}{} & P & R & F1 & P & R & F1 & P & R & F1 & F1 \\
\midrule

\multicolumn{13}{l}{\textbf{Closed (multi-turn)}} \\
\midrule
LLaMA-2 & txt       &        & 0.96 & 0.96 & 0.96 & 0.96 & 0.95 & 0.95 & 0.90 & 0.88 & 0.88 & 0.93 \\
        & txt+pr    &        & 0.95 & 0.95 & 0.95 & 0.95 & 0.95 & 0.95 & 0.94 & 0.94 & 0.94 & \textbf{0.95}**** \\
         \hline
BERT    & txt       &        & 0.95 & 0.95 & 0.95 & 0.96 & 0.96 & 0.96 & 0.95 & 0.95 & 0.95 & \textbf{0.95} \\
        & txt+pr    &        & 0.95 & 0.95 & 0.95 & 0.97 & 0.96 & 0.96 & 0.95 & 0.95 & 0.95 & \textbf{0.95} \\
         \hline
BiLSTM  & txt       &        & 0.87 & 0.87 & 0.86 & 0.90 & 0.90 & 0.90 & 0.88 & 0.88 & 0.88 & 0.88 \\
        & txt+pr    &        & 0.88 & 0.88 & 0.88 & 0.91 & 0.91 & 0.91 & 0.88 & 0.88 & 0.88 & \textbf{0.89}**** \\
         \hline
SVM     & txt       &        & 0.92 & 0.92 & 0.92 & 0.94 & 0.94 & 0.94 & 0.90 & 0.90 & 0.90 & 0.92 \\
        & txt+pr    &        & 0.93 & 0.93 & 0.93 & 0.93 & 0.93 & 0.93 & 0.92 & 0.92 & 0.92 & \textbf{0.93} \\
\midrule

\multicolumn{13}{l}{\textbf{Open (single-turn)}} \\
\midrule
LLaMA-2 & txt       &        & 0.70 & 0.51 & 0.30 & 0.78 & 0.76 & 0.77 & 0.70 & 0.61 & 0.53 & 0.57 \\
        & txt+pr    &        & 0.70 & 0.71 & 0.69 & 0.30 & 0.29 & 0.37 & 0.22 & 0.48 & 0.31 & \textbf{0.59}**** \\
        \hline
BERT    & txt       &        & 0.87 & 0.89 & 0.87 & 0.89 & 0.88 & 0.88 & 0.80 & 0.80 & 0.80 & \textbf{0.86} \\
        & txt+pr    &        & 0.91 & 0.91 & 0.91 & 0.81 & 0.81 & 0.81 & 0.72 & 0.71 & 0.69 & 0.83 \\
         \hline
BiLSTM  & txt       &        & 0.79 & 0.78 & 0.78 & 0.77 & 0.70 & 0.71 & 0.80 & 0.71 & 0.70 & \textbf{0.76} \\
        & txt+pr    &        & 0.81 & 0.76 & 0.77 & 0.80 & 0.79 & 0.80 & 0.68 & 0.68 & 0.68 & \textbf{0.76}** \\
         \hline
SVM     & txt       &        & 0.89 & 0.82 & 0.83 & 0.83 & 0.79 & 0.80 & 0.83 & 0.76 & 0.76 & 0.81 \\
        & txt+pr    &        & 0.90 & 0.86 & 0.87 & 0.87 & 0.85 & 0.85 & 0.87 & 0.84 & 0.84 & \textbf{0.86}*** \\
\bottomrule
\end{tabular}

\caption{Binary classification performance across three hate speech topics for closed (multi-turn) and open (single-turn) conversations. Models are trained on text (`txt`) or text with persuasion (`txt+pr`). We report precision (P), recall (R), and F1 scores per topic and macro-average F1. McNemar significance: ** $p<.01$, *** $p<.001$, **** $p<.0001$.}
\label{tab:resultsF1}
\end{table*}

\subsection{Classifying Hate Speech and Counterspeech Using Persuasion Modes}
\label{sec:experement_classification }
\textcolor{black}{A few studies, such as \cite{Garland2020-gv}, have tried to detect counterspeech due to the complexity of hateful conversations and the thin line between hate speech and counterspeech; indeed, in some cases, hostile and offensive language may be used in counterspeech ~\citep{Mathew2019-or,Yu2022-zb}. Thus, to answer RQ1, we designed a task to classify a given turn of a conversation as hate speech or counterspeech and evaluate the performance of incorporating a persuasion mode.} 
We experimented with the use of various model types, including the baseline model (SVM), (Bilstm), masked language model (BERT), and an instruction tuning model (Llama), on multi-turn and single-turn conversation datasets. A stratified train/validation/test split of 70\%/10\%/20\% was performed for each experiment (Appendix C provides a comprehensive overview of the dataset split details and models' main hyperparameters). 
\textcolor{black}{To incorporate the persuasion modes in a binary classification, the persuasion modes were integrated as a categorical feature. For the baseline model (SVM), we use word N-gram and Char N-gram with up to 5-grams. In the transformer model, we used BERT-based-uncased and structured a multimodal model that combines the categorical features with the textual data from the dataset. First, we created individual multilayer perceptrons (MLPs) for the categorical features (persuasion modes) and then concatenated them with the transformer output to be processed before the final classifier layer. This method can be represented as $M = T||MLP(c)$, where $M$ refers to the multimodal representation, $||$ is the concatenation operator, $T$ is the transformer output, and c represents the categorical features (persuasion mode).
For the Llama model, we incorporated specific settings into the instruction prompt to ensure the model's awareness of persuasion modes while classifying hate- and counterspeech (Appendix C).}  The instruction (prompt) tuning and parameter configuration settings in the training process for all models were the same as in the first task. We used a Bilstm model with a 64-unit bidirectional LSTM layer. This was followed by a 0.2 dropout rate, 0.2 recurrent dropout rate, and dense layer with sigmoid activation for prediction. We used the BERT-base-uncased model for BERT. The training involved 5 epochs, batch sizes of 16, and a 5e-5 learning rate. The Llama-2-7B training consisted of 20 epochs and a 2e-4 learning rate. The parameter efficient fine-tuning (PEFT) approach was employed to fine-tune a limited number of trainable parameters, reducing computational and storage expenses. For efficient fine-tuning, we used low-rank adaptation (QLoRA) to reduce the LLM memory usage without compromising performance. The results of training on text only and incorporating persuasion with the text are shown in table \ref{tab:resultsF1}. 

\subsection{Counterspeech Generation}
\label{sec:counter_gernration_experement }
\textcolor{black}{Various attempts have been made to use generative models to produce counterspeech to combat hateful comments ~\citep{Sen2023-le} or to rephrase hateful comments and mitigate the messages' offensiveness~\citep{Agarwal2023-lo}. Thus, in this study, to further examine the efficacy of the machine-generated counterspeech, we evaluated persuasion modes in it, compared them to human persuasion modes, and analyzed language models' ability to generate counterspeech.} We employed Chat GPT-3 and Llama 2 (the most common instruction-based language models) to generate counterspeech for each hateful comment in multi-turn and single-turn conversations. Primarily, we used the open-source Llama 2 Chat 13B~\citep{Touvron2023-my} model, which was fine-tuned on the publicly available instruction datasets with 13 billion parameters using the GGML format model (a library for machine learning that allows CPU inference). We also used the gpt-3.5-turbo-0613 model accessed via the OpenAI API\footnote{\url{https://openai.com/blog/gpt-3-5-turbo-fine-tuning-and-api-updates}}. We instructed the models using the direct method without few-shot fine-tuning in the counterspeech data since the aim was to assess the persuasion modes that the models use (i.e., zero-shot setting with no examples provided). We used the same instructions for the two models to force counterspeech generation. The model GPT-3.5 tends to refuse to generate counterspeech for offensive posts when directly prompted. We used a task-specific prompt indicating that the model needed to generate a counterargument and explained the task in the instructions. We used both models' standard system prompt styles and the same instructions to indicate that the task was to generate counterspeech.\footnote{Appendix D provides a sample of the instructions for the zero-shot instruction model to generate counterspeech for hateful posts.}

\subsubsection{Machine-Generated Counterspeech Labeling}
\label{sec:counter_gernration_labeling }
After generating counterspeech for the datasets, a multi-classification approach was applied to the gold persuasion labels to populate the weakly supervised labels of persuasion modes in the machine-generated counterspeech models. This weak labeling approach was adopted to label the output of the generated text to maintain the annotation cost, especially if the annotation needed to be done for multiple generative model outputs ~\citep{Sen2023-le}. More specifically, for persuasion labeling in the conversations, weak supervision modeling of persuasion has been shown to be effective in previous studies ~\citep{Chen2021-mw}. In their previous study~\citep{Chen2021-mw} used a weakly supervised model that utilizes partially labeled data to predict associated persuasive strategies for each sentence. Thus, to maintain the cost of annotation, in this study, we experimented with different models: baseline (SVM), masked language model (BERT), and instruction-tuning model (Llama 2). We trained these models to predict the persuasion modes based on the gold annotations from the two datasets using the same split of training and testing reported in Appendix C. The result of training the models to predict the persuasion modes and the hyperparameters used in this task were consistent with those employed in the binary classification task, as further detailed in Appendix D. 
\begin{figure*}[!ht]
  \centering

  \begin{subfigure}[b]{0.48\textwidth}
    \centering
    \includegraphics[width=\linewidth]{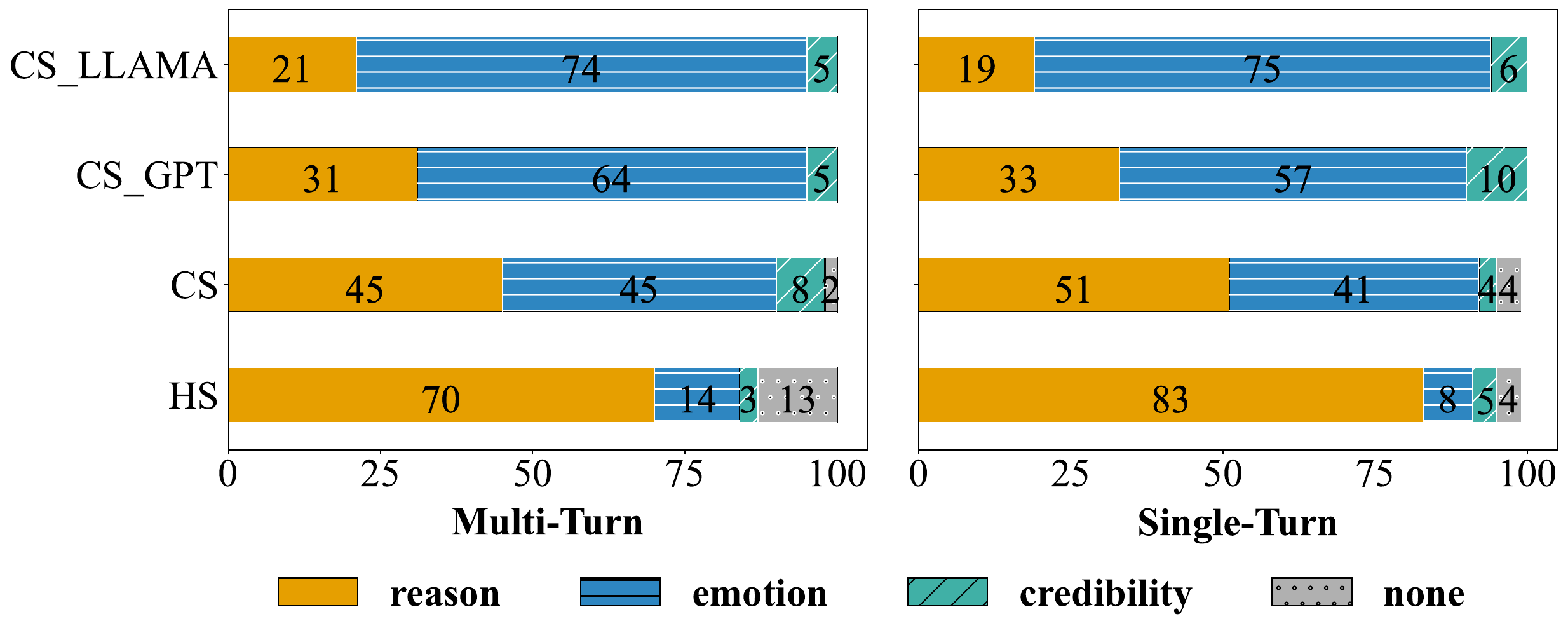}
    \caption{Overall persuasion mode distribution}
    \label{fig:overall}
  \end{subfigure}
  \hfill
  \begin{subfigure}[b]{0.48\textwidth}
    \centering
    \includegraphics[width=\linewidth]{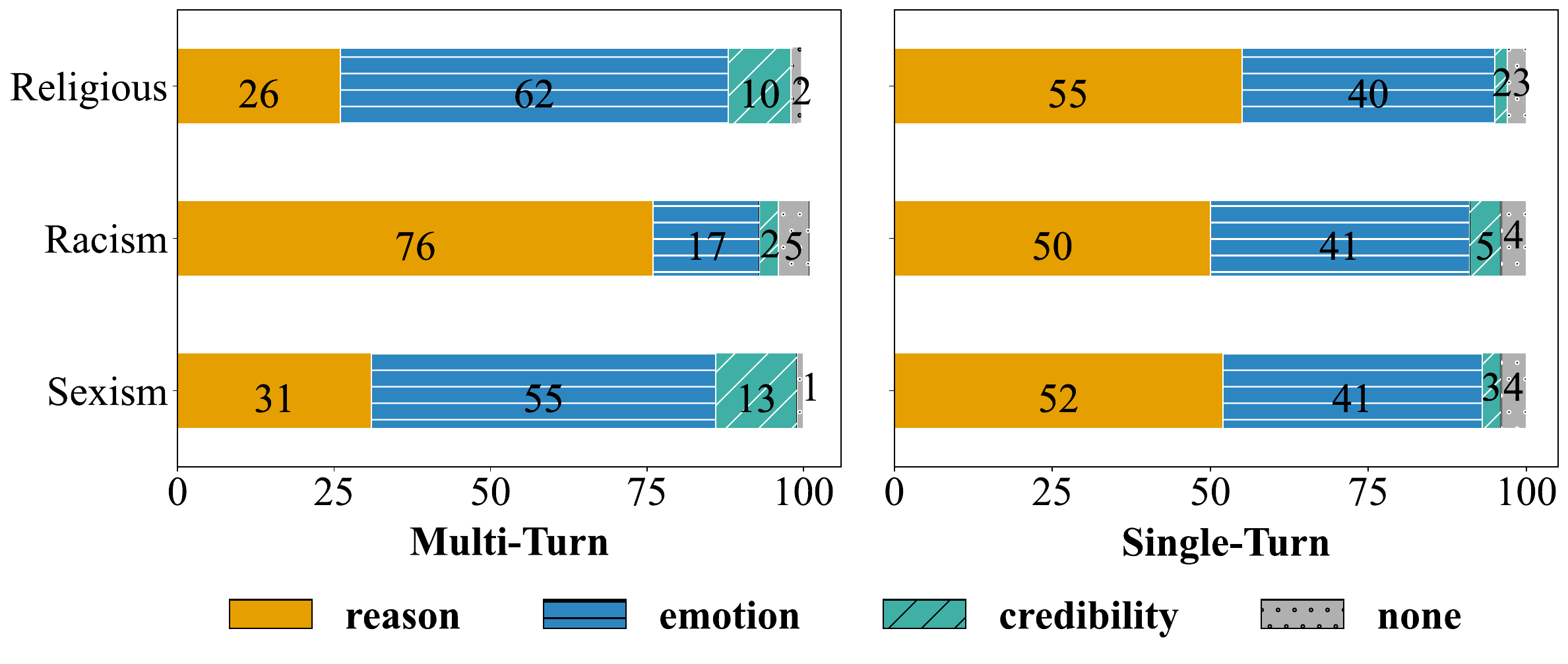}
    \caption{Human counterspeech by topic}
    \label{fig:dist_human_topic}
  \end{subfigure}


  \begin{subfigure}[b]{0.48\textwidth}
    \centering
    \includegraphics[width=\linewidth]{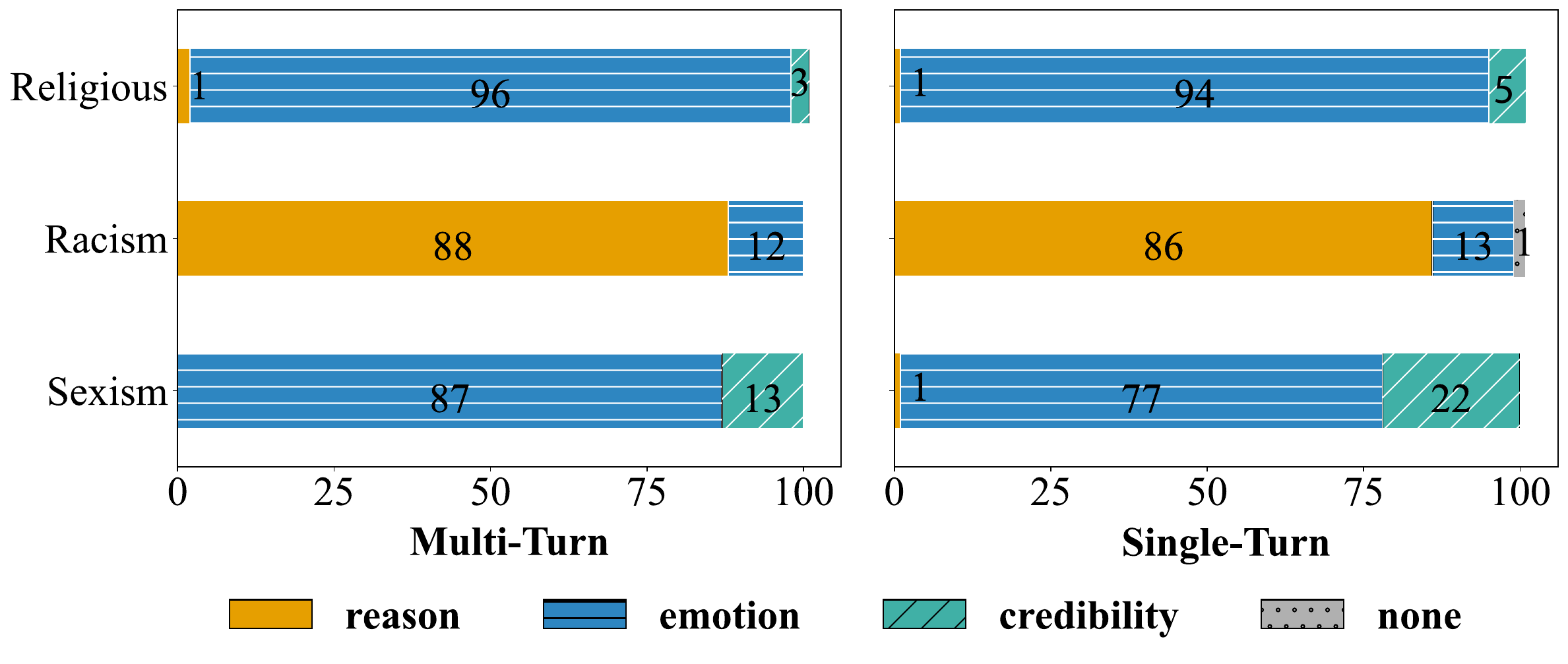}
    \caption{GPT-3 counterspeech by topic}
    \label{fig:dist_gpt_topic}
  \end{subfigure}
  \hfill
  \begin{subfigure}[b]{0.48\textwidth}
    \centering
    \includegraphics[width=\linewidth]{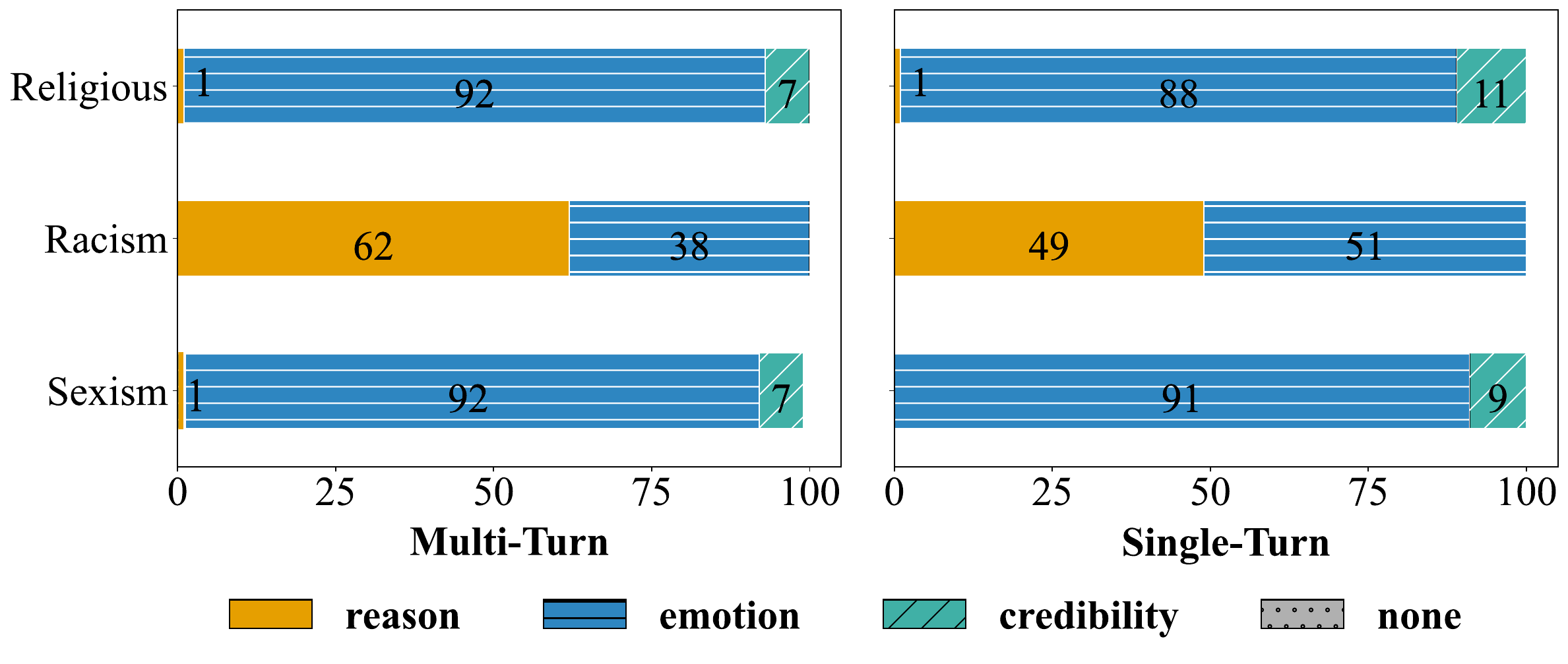}
    \caption{Llama 2 counterspeech by topic}
    \label{fig:dist_llama_topic}
  \end{subfigure}

  \caption{Distribution of persuasion modes (in \%) in counterspeech across overall and topic-specific contexts, comparing human, GPT-3, and LLaMA 2 responses in multi-turn (closed) and single-turn (open) interactions.}
  \label{fig:dist_all_2x2}
\end{figure*}

\subsubsection{Validation of Counter Generative Persuasion Labels}
\label{sec:validation_counter_gernration_experement }
We ran both benchmark and human verification validations. \textbf{In the benchmark validation}, the remaining closed DialogConan dataset, which was not included in our gold annotations, was used to verify the general model behavior. First, we compared the results of predicting the persuasion classes in the silver (weak label) and gold label benchmarks. We detail the process in Appendix D and show that the behavior of the models trained on silver labels is consistent with the models trained on gold labels. This consistent pattern across diverse models reinforces the weakly supervised model's efficacy in predicting the persuasion modes in the machine-generated counterspeech dataset. 

\textbf{In the human validation}, we manually examined the first 50 examples for each topic. To validate the result of the weakly supervised labels for the generated counter, we selected 50 generated counters from each topic and generative model and validated the persuasion label. An independent annotator (one of the authors) performed the validation process through the two datasets to ensure the consistency of the labeling process. In general, in the closed dataset, the overall error rate is relatively small: about 19\% for GPT-3 and 12\% for Llama 2. Similarly, for the open dataset, the overall error rate is about 22\% for GPT-3 and 16\% for Llama 2 (more details in Appendix D). Additionally, we investigated the significance of the comparison between the persuasion labels of humans and generated a counter using the chi-squared $X^2$, as shown in Table ~\ref{tab:significant_generative}, which shows that all the comparisons are statistically significant.

\begin{table} [!h]

    \centering
 
\resizebox{\linewidth}{!}{%
    \begin{tabular}{c|c|c}
        Topic-Type & Human-GPT
 & Human-LLAMA
 \\
        \hline
        Religious bigotry (mT) & 1.08e-101 **** & 3.55e-90 ****\\
        Racism (mT) &  1.22e-15 **** & 1.04e-37 ****\\
        Sexism (mT) & 5.95e-113 **** &  1.32e-124 ****\\
        Religious bigotry (1T) & 4.74e-22 ****& 1.72e-22 ****\\
        Sexism (1T) & 4.81e-40 **** & 1.88e-38 ****\\
        Racism (1T) & 1.08e-14 ****&0.007 ***\\
    \end{tabular}}
    \caption{Comparison of generative and human counterspeech for multi-turn (mT) and single-turn (1T). Using the chi-squared $X^2$ significant test, all comparisons are statistically significant 
(p $<$ 1.00), ** (p $<$ .01), *** (p $<$ .001), and **** (p $<$ .0001)}
    \label{tab:significant_generative}
\end{table}

\section{Results and Analysis}
\label{sec:results }
In this section, we present the results of the previous experiments and interpret the findings related to each research question. 

\subsubsection{Interplay of Persuasion Modes in Hate Speech and Counterspeech in Each Type of Conversation Interaction, RQ1} 
We trained various models to examine the effectiveness of persuasion modes in identifying hate speech and counterspeech. Mainly, we evaluated the general capability of the persuasion modes as proxies to predict hateful and counterspeech responses in two datasets open and closed conversation interactions.
Table ~\ref{tab:resultsF1} illustrates a slight enhancement in the models' performance when the persuasion modes are incorporated as categorical features. Some models (such as the masked language model BERT) have more resistance and an overall stable performance in closed interactions, with a small increase in the topic racism's precision score (approximately a 97\% F1 score). The same model behavior can be observed in open interactions, in which BERT performs better when text-only input is used instead of text and persuasion on the overall average macro (F1 score). However, for topics such as racism, the combination of text and persuasion has a relative enhancement in the F1 scores, Bilstm with a 91\% F1 score, while the text-only F1 score is approximately 90\%. The same behavior can be observed with open interactions in BiLSTM, in which the performances of the text-only input and the input combination of text and persuasion are stable, with F1 scores of approximately 76\%. \textcolor{black}{This stable performance of BERT may be partly due to the pre-trained language models' data shift effect, as some topics have better representation than others either as a result of covariate shift~\citep{Broscheit2022-zj} or temporal domain adaptation~\citep{Rottger2021-zy}. For example, the study by ~\citep{Florio2020-ux} shows that hate speech detection models trained on data temporally closer to the test data perform better with transformer-based models.}
Relatively less open interaction data are present at the topic level, table ~\ref{tab:datasets_hs_cs_dis}. For instance, the topic of religious bigotry has 279 samples, and the prediction score on this topic has the lowest F1 performance in BiLSTM. This model behavior also can be confirmed with SVM open interaction performance because it has a better F1 score than the instruction-tuning model Llama, with an F1 score of 86\%. However, in the closed dataset, BiLSTM improves slightly when persuasion is incorporated with the text, with an overall F1 score of 89\% compared to the F1 score of 88\% obtained when the text is used alone. Moreover, the instruction-tuning model's (Llama 2) performance is enhanced when persuasion modes are incorporated as a feature, as shown by a few examples of instruction fine-tuning (an overall F1 score of 95\% in closed [multi-turn] dataset). This can also be seen in the open dataset used in Llama 2, which improved when it used text and persuasion (59\% F1) compared to text alone (57\% F1). \textcolor{black}{It is worth mentioning the discrepancies in the performance of Llama 2 when fine-tuned on open versus closed dataset (overall 95\% F1 score in closed dataset, and only 59\% for open dataset). We can link the poor performance of Llama 2 fine-tuned on the open dataset to the possible data contamination of benchmark datasets ~\citep{Sainz2023-ne}. We must consider that the closed conversation dataset is a well-known benchmark that contains a seed dataset from the CONAN dialogues released mainly in 2021 and extended to the final set in 2022 ~\citep{Bonaldi2022-yj}. We must also consider that the knowledge cutoff month for Llama 2 series models is September 2022 according to the model card ~\citep{Touvron2023-my}. As ~\citep{Bonaldi2022-yj} indicate, a major set of the DilogConan closed dataset that we used in this study was derived from using part of the data presented in ~\citep{Fanton2021-bc}'s study as an additional resource that consists of 5,000 hate speech – counterspeech pairs. Data contamination is a major issue in pre-trained language models, as a recent study by ~\citep{Jiang2024-rc} found evidence that some LLMs have seen task examples during pre-training for a range of tasks and are therefore no longer zero- or few-shot for these tasks. Another possible reason is the noisy nature of the open conversation dataset, as it has been derived from social media platforms ~\citep{Albanyan2023-fc}. The open interaction conversation dataset has the distinct feature of enabling the broader public (many users) to engage with a post compared to closed interaction conversations in which the dialog is exclusively between two users, one of whom may be an expert annotator (mitigation procedure expert)~\cite{Fanton2021-bc}.}

\begin{figure*}[!h]
\centering

\makebox[0.45\textwidth][c]{\textbf{Multi-turn}}%
\hfill
\makebox[0.45\textwidth][c]{\textbf{Single-turn}}


\begin{subfigure}[b]{0.45\textwidth}
  \includegraphics[width=\linewidth]{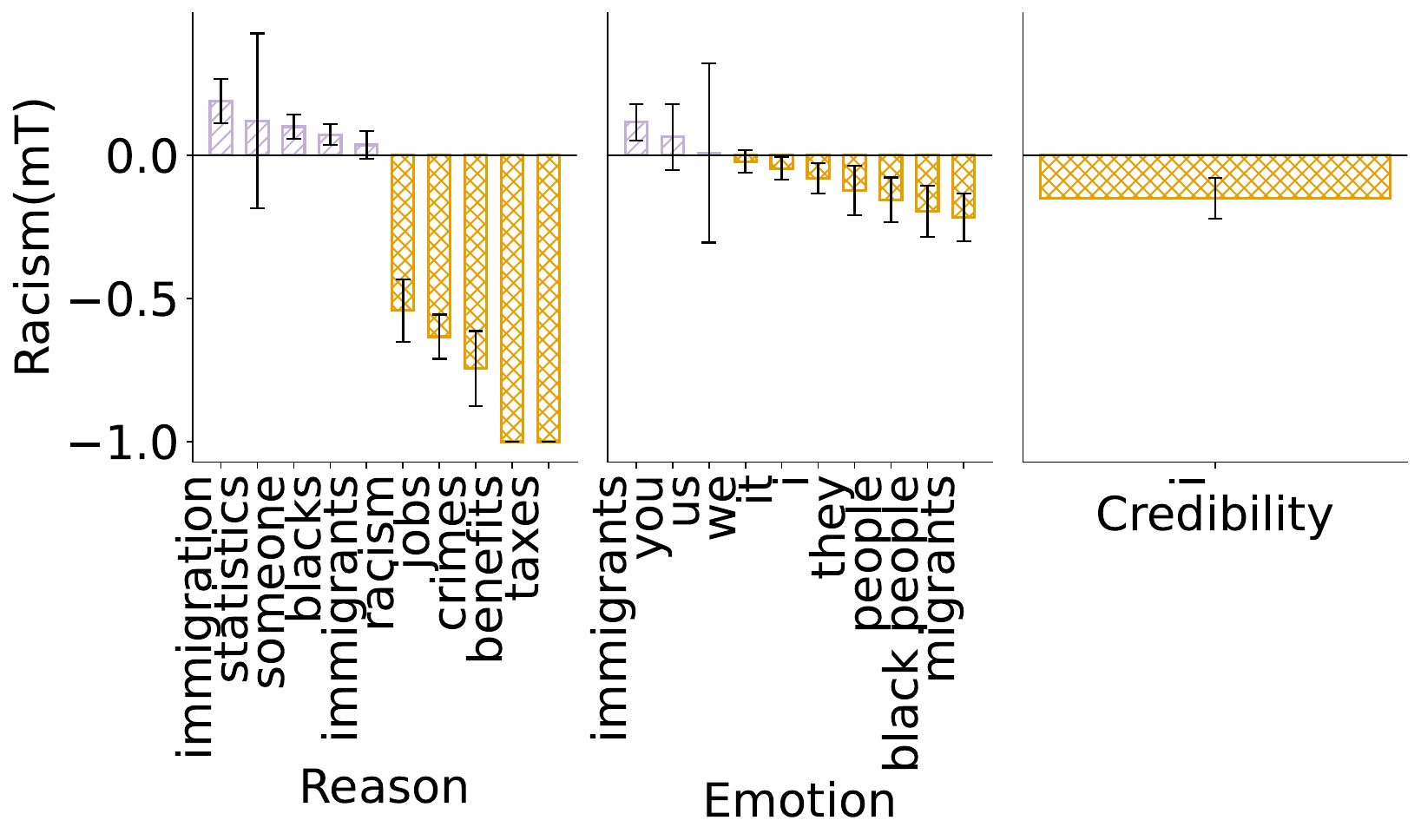}
  \caption{Racism (Multi-turn)}
  \label{fig:multi_Racism}
\end{subfigure}
\hfill
\begin{subfigure}[b]{0.45\textwidth}
  \includegraphics[width=\linewidth]{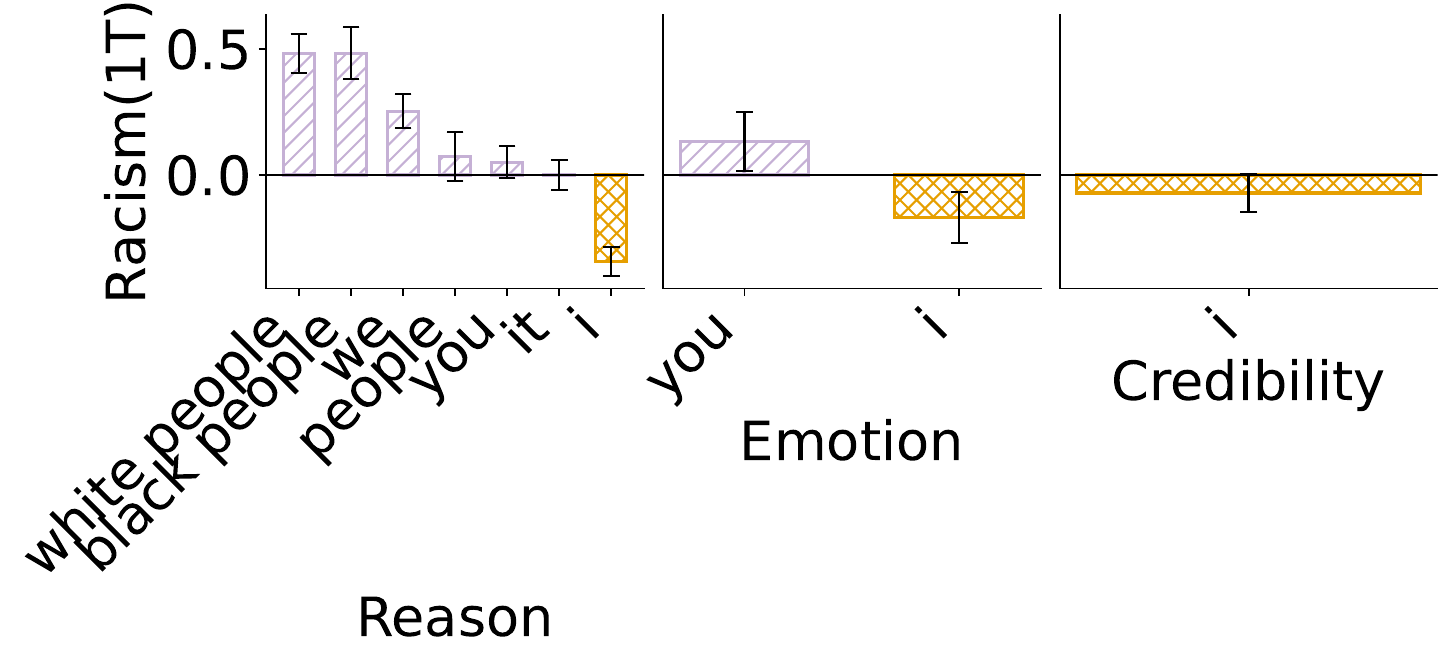}
  \caption{Racism (Single-turn)}
  \label{fig:single_racism}
\end{subfigure}


\begin{subfigure}[b]{0.45\textwidth}
  \includegraphics[width=\linewidth]{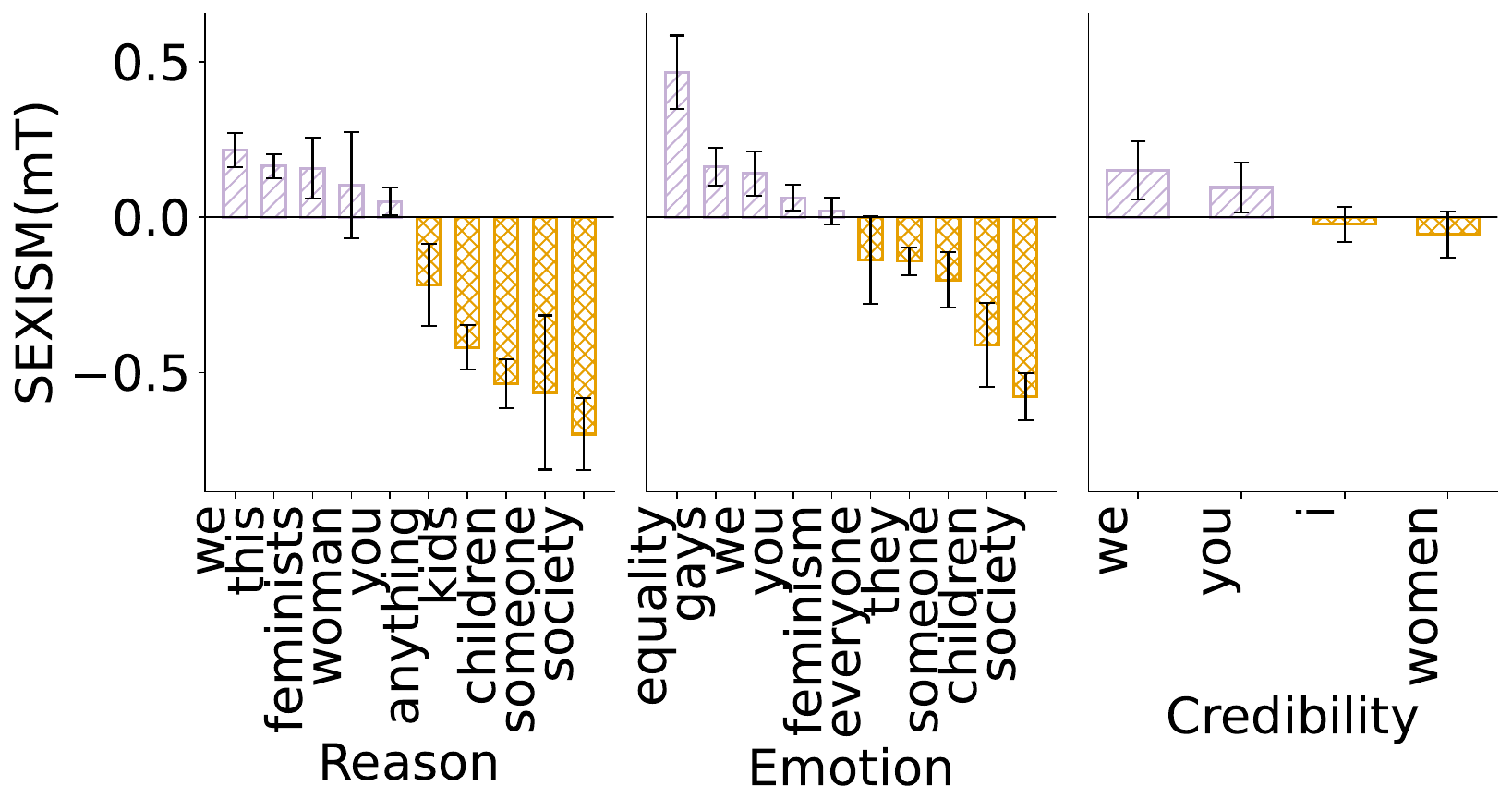}
  \caption{Sexism (Multi-turn)}
  \label{fig:multi_Sexism}
\end{subfigure}
\hfill
\begin{subfigure}[b]{0.45\textwidth}
  \includegraphics[width=\linewidth]{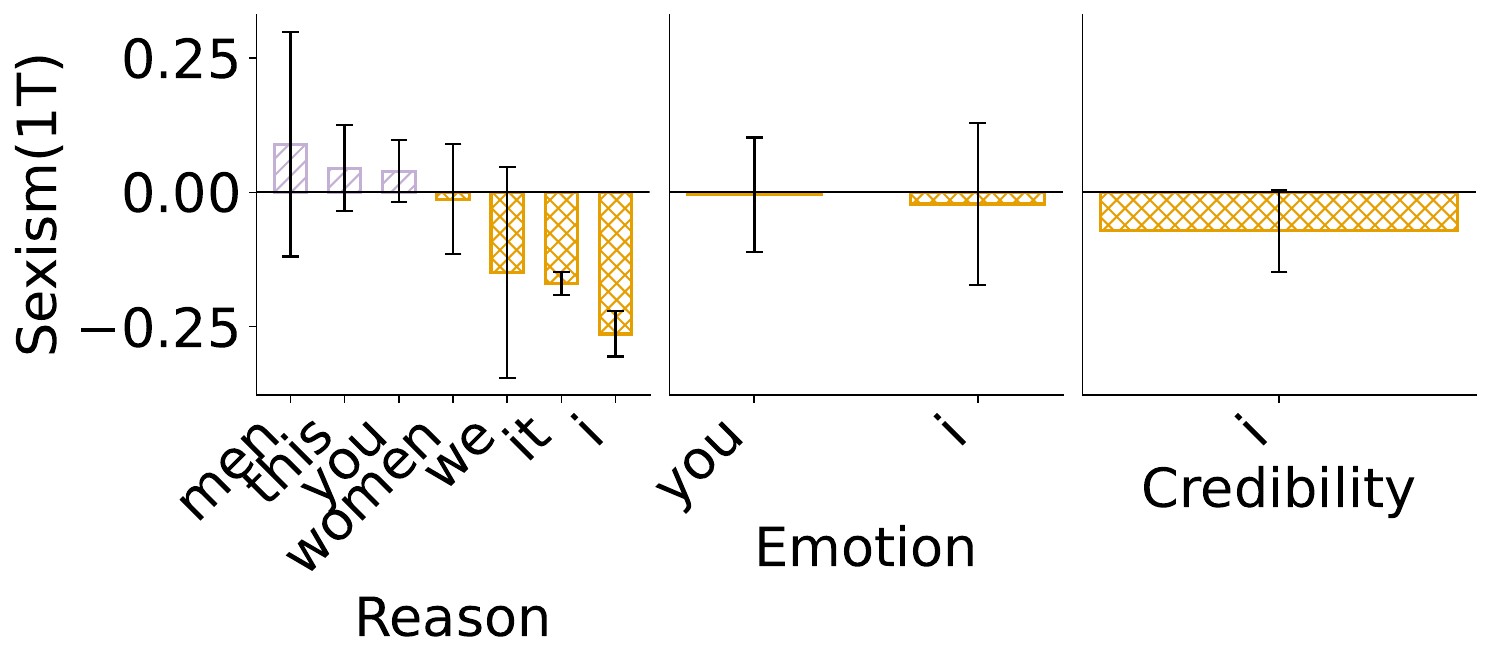}
  \caption{Sexism (Single-turn)}
  \label{fig:single_sexism}
\end{subfigure}


\begin{subfigure}[b]{0.45\textwidth}
  \includegraphics[width=0.85\linewidth]{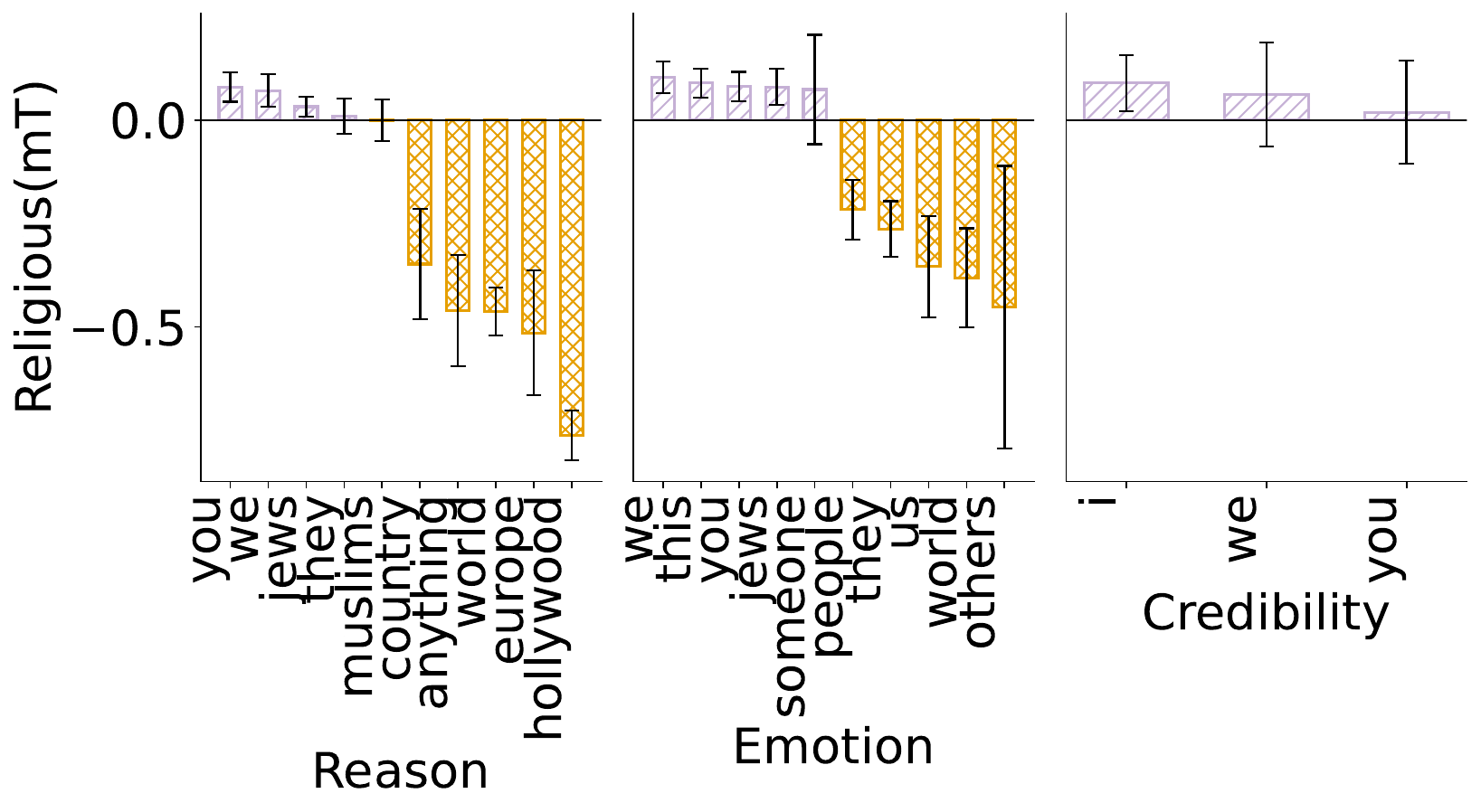}
  \caption{Religious bigotry (Multi-turn)}
  \label{fig:multi_Religious}
\end{subfigure}
\hfill
\begin{subfigure}[b]{0.45\textwidth}
  \includegraphics[width=\linewidth]{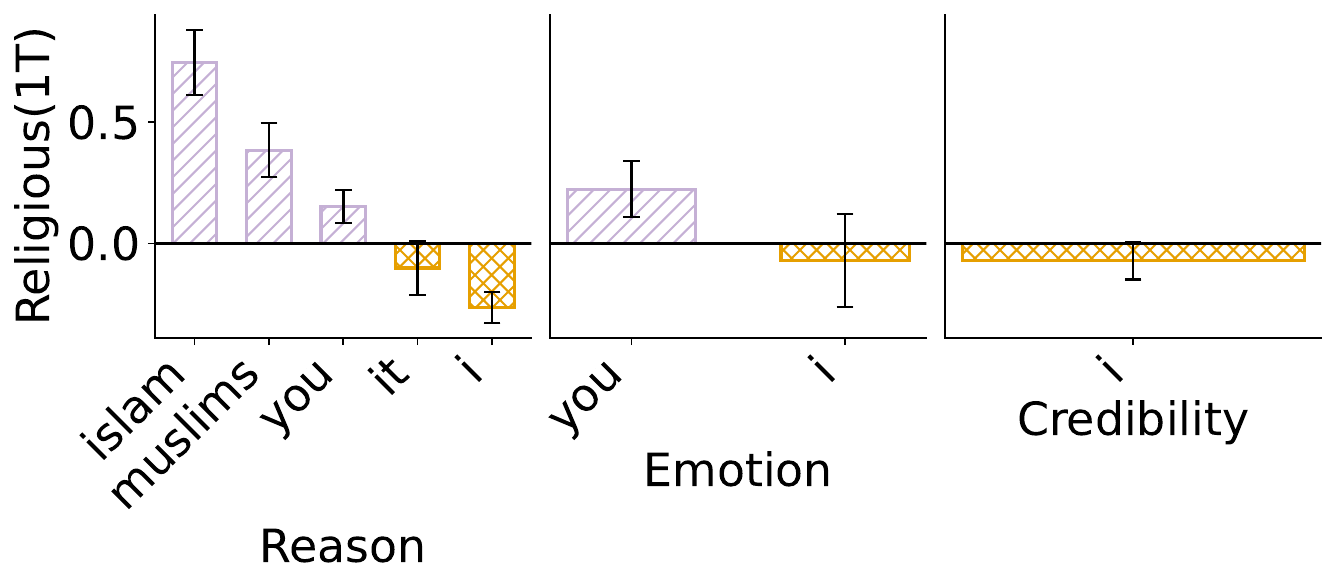}
  \caption{Religious bigotry (Single-turn)}
  \label{fig:single_religion}
\end{subfigure}


\caption{
Entities identified by Riveter and power scores of personas across multi-turn (mT) and single-turn (1T) conversations for three topics (racism, sexism, and religious bigotry). A positive value indicates a stronger association with the persuasion mode. 
\includegraphics[height=1em]{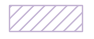} Positive, 
\includegraphics[height=1em]{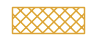} Negative.
}
\label{fig:combined_turns}
\end{figure*}

\subsubsection{Persuasion mode variations expressed in generated and human counterspeech, RQ2} 
We analyzed the persuasion mode distribution by conversation interaction type and generative- and human-based counterspeech (see Figure~\ref{fig:overall}). \textcolor{black} {We also used the entity recognition and coreference resolution capabilities of Riveter ~\citep{Antoniak2023-bt} to identify which characters are framed as having or lacking power and assess the social dynamics of personae in the conversations for each persuasion type (Figures~\ref{fig:multi_Racism}-~\ref{fig:single_religion}). }
Overall, generative counterspeech persuades using emotions, while humans persuade using reason. As shown in Figure~\ref{fig:overall}, for closed (multi-turn) interactions, the reason persuasion mode comprises approximately 45\% of human counterspeech, compared to only 31\% and 21\% of generative counterspeech (GPT and Llama-2, respectively). Similarly, in open interactions, 51\% of human counterspeech uses reason, compared with 33\% and 19\% in the GPT and Llama-2 generative models, respectively.   
Looking closely at the topic level, the reason persuasion mode is more often used in racism than in religious bigotry and sexism. Emotion, however, is more often used in counterspeech, especially in closed (multi-turn) interactions Figure~\ref{fig:dist_human_topic}. This finding can be further reiterated by reviewing the persona dynamics in the racism topic using Riveter (Figures~\ref{fig:multi_Racism} -~\ref{fig:single_racism}). Using reasoning statements that contain ''immigration'' and ''statistics'' as entities have more powerful persona scores in conversations when discussing rationales to support counterpoints in closed (multi-turn) interactions (see Figure~\ref{fig:multi_Racism}). These persona tend to be fact-based rationale (immigration rules and statistical studies), while persuasion in open interactions tends to use personal characteristics (e.g., skin color). Conversely, in conversations that used emotion-based persuasion, the pronoun "you" had the highest persona dynamic power score. This mode in conversations tends to reflect shameful feelings shown by posters of hate speech. Interestingly, credibility has a slightly higher presence in human-authored counterspeech (8\%) compared to machine-generated counterspeech (5\%; Figure~\ref{fig:overall}). As in closed (multi-turn) interactions, human-authored counterspeech supports the argument by using personal stories. For instance, on the topic of religious bigotry, conversations using the credibility persuasion mode have personal perspectives (with the first-person pronoun "I") when personal stories are recounted (Figure ~\ref{fig:multi_Religious}). On the topics of religious bigotry and sexism, conversations use credibility more than they do in the topic of racism, in both human-sourced and machine-generated counterspeech.
 \begin{figure*}[t!]
    \centering
    \begin{minipage}[t]{0.63\textwidth}
        \centering
        \includegraphics[width=\linewidth]{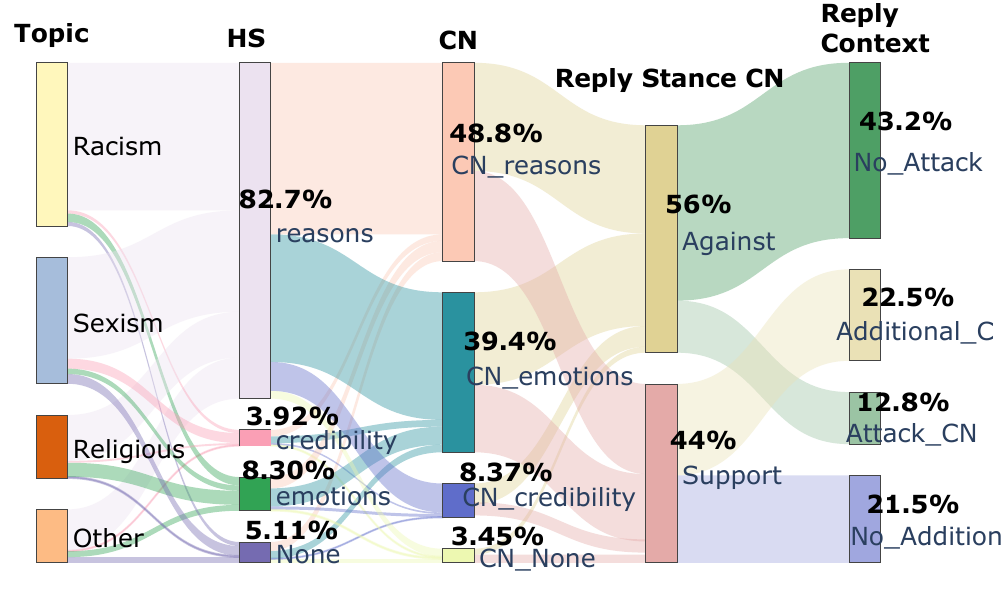}
        \caption{Flow of persuasion modes between hate speech (HS) and counternarratives (CNs) in single-turn conversations (open interactions).}
        \label{fig:sanky}
    \end{minipage}
    \hfill
    \begin{minipage}[t]{0.32\textwidth}
        \centering
        \includegraphics[width=\linewidth]{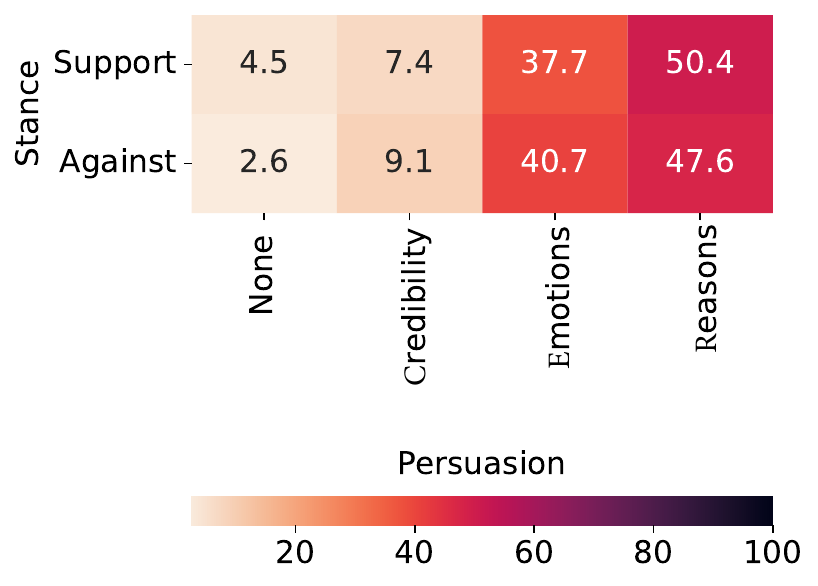}
        \caption{Correlation between the persuasion modes and the reply's stance.}
        \label{fig:heatmap}
    \end{minipage}
\end{figure*}

\subsubsection{The reply's stance toward counterspeech, RQ3}
 \textcolor{black}{The open interaction conversation data has the distinct feature of enabling the broader public (many users) to engage with the post compared to closed interaction conversations in which the dialog is exclusively between two users (such as on GPT and chat platforms). We use open interaction conversation data to examine the stance of the replies to counterspeech. In contrast to previous studies, such as ~\citep{Mathew2019-or}, which considered the presence of replies and likes on a post as indicative of support to the counterspeech, our study implemented the nuance of assessing the element in counterspeech that makes it likely to receive support by focusing on the persuasion mode.} Figure ~\ref{fig:sanky} illustrates the conversation flow between hate speech and counterspeech, as well as the stance of the replies received by the counterspeech. The general trend observed is that counterspeech receives replies that are against it (about 56\%). Moreover, supportive replies tend to add context to further justify the counterpoint (22.5\%). Most of the opposing replies (43\%) do not directly attack the counterspeech; instead, they reject the counterpoint. We examined the distribution of supportive and opposing stances using the persuasion mode and present this in Figure~\ref{fig:heatmap}. There is a noticeable tendency for replies to support counterspeech that uses the reason persuasion mode (50.4\%), and replies against the counterspeech are at 74\%. Replies that use the emotion persuasion mode and are opposed to the counterspeech are at 40.7\%, and those that use the same persuasion mode and are supportive of the counterspeech are at 37.7\%.

\section{Discussion and Implication}
\label{sec:discussion}
In this section, we interpret the results of our analysis by first discussing the interplay between the persuasion mode and counterspeech. Following this, we clarify the multifaceted impact of the interaction type and propose possible directions for future work.
The results show that the persuasion mode can provide an explainability method to counterspeech studies.
\\

\subsubsection{Conversation Type and Persuasion Modes} 
Interaction types of the conversation clearly have nuanced effects on counterspeech's exhibited persuasion mode. As shown in Figure \ref{fig:overall}, 51\% of the counterspeech examples in open interactions use reason as the persuasion mode, while 45\% of the counterspeech examples in closed interactions use reason. This may be partly due to the cost of public opinion and the users' 
need to appear knowledgeable, which may drive their tendency to support their counterpoints with facts and reasoning. \textcolor{black}{By studying the persona's social dynamic in hate conversations using Riveter ~\citep{Antoniak2023-bt}, as shown in (Figures ~\ref{fig:multi_Racism}–~\ref{fig:single_religion}),} we can see this behavior, its high power score toward people of a targeted group (White or Black people), and its lower power score when using self-reference (I). 
This persuasion mode tends to make the the conversation focus on justifying counterpoints to the target group rather than expressing the reason behind the hate and linking it to the self as a valid reason for hate. We also show the effectiveness of using persuasion as a proxy to differentiate counterspeech from hate speech (Table~\ref{tab:resultsF1}), which highlights the persuasion mode's ability to serve as an instrument of explanation for hate detection model interpretability. Overall, these results offer an overview of the counterspeech's nuances and the conversation interaction type's effect on how those confronting hate may incorporate persuasion modes to express their counterpoint and combat hate.

\subsubsection{Generative Counterspeech Persuading Using Emotions, while Humans Persuading with Reason} 
Our results show a clear gap in the singular focus of current counterspeech works on emotion-based hate-speech analyses, which further confirms other works' recent findings~\citep{Lahnala2022-oa}. For instance, the work by~\citep{Lahnala2022-oa} highlights the need to look beyond emotion, as it shows that cognitive empathy mitigates hate better than emotional empathy. Figure ~\ref{fig:overall} shows the clear tendency of human-written counterspeech to use reason in order to express a counterpoint (over 45\%); this is comparable to the mere 19\% of machine-generated counterspeech that uses reason. Further, emotion is the main persuasion mode. Generative counterspeech's behavior highlights the vicious cycle of training hate speech detection models on emotion-based data and the complexity of incorporating reason as a factor in the current generated content. A few recent studies have highlighted machine-generated counterspeech's limitations (e.g., ~\citep{Mun2023-gv,Sen2023-le}). Their conclusion focuses on a target-specific strategy and shows a lack of persuasive power in generated counterspeech when compared to human-written counterspeech. Our study's findings intensify the call for studying counterspeech's non-emotional factors as building blocks of hate mitigation.

\subsubsection{Conversation Structures and Reply Stance} 
Open interaction conversations (such as those on social media) are a means to observe how the public perceives counterspeech from multiple users' perspectives. Thus, we analyzed the replies to counterspeech and showed the stances of the replies toward each type of persuasion mode. The results in Figure~\ref{fig:sanky} show that the conversational flow alternates between the counterspeech and the type of the reply's stance. Approximately 59\% of counterspeech examples received replies with opposing stances, especially if the examples used the emotional persuasion mode (representing about 40.7\% of the opposing stances, as in Figure ~\ref{fig:heatmap}). A reply's support stance is likely to favor the reason persuasion mode, as represented in about 50.4\% of supportive replies. Since this study's scope is to provide a high-level comparison of the conversation interaction types and persuasion mode, possible future research may incorporate a nuanced analysis of how users perceive counterspeech by a specific topic. Recent studies have investigated the role of conversation structure while studying hate detection. For instance, the work by ~\citep{Yi2023-lz} surveyed session-based cyberbullying and discussed the structure of social media conversations to construct a cyberbullying dataset. Our study shows an empirical investigation that paves the way for future consideration of replies to counterspeech to enhance the interpretation of counterspeech effective factors. The size of the open interaction has its limitations in this study. \textcolor{black}{Moreover, we justified the results using chi-squared $X^2$ significant test to validate the interpretation of the comparison between machine-generated and human-authored counterspeech (Table ~\ref{tab:significant_generative}). }
\section{Conclusion}
\label{sec:conclusion }
In this study, we examined persuasion modes in two types of conversations: closed (multi-turn) and open (single-turn) interactions. First, we categorized three types of persuasion in counterspeech: emotion, reason, and credibility. Next, we assessed the persuasion modes in machine-generated counterspeech and compared them to human-authored counterspeech. The findings mainly demonstrate that the persuasion modes used in counterspeech is nuanced and noticeably dependent on the conversation topic. Emotion as a persuasion mode is more prevalently used in machine-generated counterspeech than in human-composed counterspeech, which primarily uses reason to defend counterpoints. Reason-based counterspeech tends to obtain more supportive responses compared to emotion-based counterspeech. 
Further, our work highlights the potential of incorporating persuasion modes in studies on countering hate speech to serve as explanatory proxies to further enhance the ability to interpret hate speech detection. Current counterspeech studies lack a clear method with which to evaluate optimal counterspeech strategies. The conversation structure in hate-speech detection needs to consider including the replies' stances on counterspeech as analysis factors. This feature has a clear presence in open interactions, in which multiple users can reply to counterspeech. This study focused on hate speech as a general category. Future research may further evaluate the persuasion modes using more fine-grained base types of hate (e.g., offensiveness, toxicity, or abusiveness). A narrower domain-level counterspeech evaluation can be performed by studying the topic's social aspects, such as racial discrimination against individuals or groups.

\section*{Ethical Statement}
\label{sec:ethical }
In this study, we extended current research attempts to examine the effective factors of counterspeech as a step to mitigate hate speech. Yet, we have cautiously handled the extra annotation layer (persuasion mode) that we added to the existing two released datasets by ensuring the anonymity of the posts using the IDs in the X posts. Additionally, we carefully handled the data annotation process by removing any mention of the user account from the open interaction dataset, ContextCounter~\citep{Albanyan2023-fc}. We also ensured a fair wage for the annotators by using the LabelBox annotation platform's fixed pricing per hour (as explained in Appendix B). Moreover, we tried to limit the direct reporting of hate messages as a sample for analysis by using Riveter ~\citep{Antoniak2023-bt} to measure the power and dynamics between the social entity of conversations and help with interpreting the narrative of hate speech and counterspeech. 



\footnotesize
\bibliography{Hate} 
\subsection*{Paper Checklist}

\begin{enumerate}

\item For most authors...
\begin{enumerate}
    \item  Would answering this research question advance science without violating social contracts, such as violating privacy norms, perpetuating unfair profiling, exacerbating the socio-economic divide, or implying disrespect to societies or cultures?
    \answerYes{yes, we explain the motivation at \cref{sec:intro}, and detailed the research implication at \cref{sec:discussion} }
  \item Do your main claims in the abstract and introduction accurately reflect the paper's contributions and scope?
    \answerYes{yes, we detail the study contributions at \cref{sec:intro}, and provide detailed discussion of the results for each research question at \cref{sec:results }}
   \item Do you clarify how the proposed methodological approach is appropriate for the claims made? 
    \answerYes{yes, we start by providing the definition of counter speech and the hypothesis of persuasion mode at\cref{sec:intro} and situate the contribution with the related work at \cref{sec:previous_work} RQ1 indicated at mainly at the methodology \cref{sec:experement_classification }, and at \cref{sec:intro}. As for RQ2, we compare the different persuasion modes in humans and generated counters using the distribution of the persuasion in two types of data: closed (multi-turn) and open (single-turn) as explained at \cref{sec:counter_gernration_experement }, \cref{sec:intro} and \cref{sec:results }. For RQ3, we explain the flow of replies at \cref{sec:dataset_Experemnt}, and further details at \cref{sec:discussion} }
   \item Do you clarify what are possible artifacts in the data used, given population-specific distributions?
    \answerYes{yes, we report the detailed data selection and preparation at \cref{sec:dataset_Experemnt} and Appendix A.}
  \item Did you describe the limitations of your work?
    \answerYes{yes, at discussion section\cref{sec:discussion}}
  \item Did you discuss any potential negative societal impacts of your work?
    \answerYes{We detail the research implication at \cref{sec:discussion} and ethical statement \cref{sec:ethical } }
      \item Did you discuss any potential misuse of your work?
    \answerYes{yes, discussion of research implication at \cref{sec:discussion}and ethical statement \cref{sec:ethical }}
    \item Did you describe steps taken to prevent or mitigate potential negative outcomes of the research, such as data and model documentation, data anonymization, responsible release, access control, and the reproducibility of findings?
    \answerYes{yes, at ethical statement \cref{sec:ethical }, and appendix A.}
  \item Have you read the ethics review guidelines and ensured that your paper conforms to them?
    \answerYes{yes}
\end{enumerate}

\item Additionally, if your study involves hypotheses testing...
\begin{enumerate}
  \item Did you clearly state the assumptions underlying all theoretical results?
    \answerYes{We provide empirical justification to the hypothesis and answer the research question.}
  \item Have you provided justifications for all theoretical results?
    \answerYes{We provide empirical justification to the hypothesis and answer the research question.}
  \item Did you discuss competing hypotheses or theories that might challenge or complement your theoretical results?
    \answerYes{We describe the main hypothesis at \cref{sec:intro} and situate the assumption with previous work \cref{sec:previous_work}}
  \item Have you considered alternative mechanisms or explanations that might account for the same outcomes observed in your study?
    \answerYes{yes, full analysis of the results described at \cref{sec:results }, and we provide further details at appendix D for generated counter analysis.}
  \item Did you address potential biases or limitations in your theoretical framework?
    \answerYes{yes, we detail the discussion at \cref{sec:discussion}}
  \item Have you related your theoretical results to the existing literature in social science?
    \answerYes{yes, \cref{sec:previous_work}, and research implication at \cref{sec:discussion}}
  \item Did you discuss the implications of your theoretical results for policy, practice, or further research in the social science domain?
    \answerYes{yes, we show in the discussion section a detailed implication of the empirical results \cref{sec:discussion}}
\end{enumerate}

\item Additionally, if you are including theoretical proofs...
\begin{enumerate}
  \item Did you state the full set of assumptions of all theoretical results?
    \answerYes{yes, we explain the hypothesis of the experiment design at \cref{sec:intro} and further justify it by sitting it with previous work \cref{sec:previous_work}}
	\item Did you include complete proofs of all theoretical results?
    \answerYes{yes, full experimental results and discussion at sections \cref{sec:results } and \cref{sec:discussion}}
\end{enumerate}

\item Additionally, if you ran machine learning experiments...
\begin{enumerate}
  \item Did you include the code, data, and instructions needed to reproduce the main experimental results (either in the supplemental material or as a URL)?
    \answerYes{detail explanation of the experiment hyperparameter or data annotation guideline are at Appendix A, B, C, and D.}
  \item Did you specify all the training details (e.g., data splits, hyperparameters, how they were chosen)?
    \answerYes{yes, we provide a detailed description of the training split at appendix C. }
     \item Did you report error bars (e.g., with respect to the random seed after running experiments multiple times)?
    \answerYes{yes, full detail of significant text and experiment design at appendix C. }
	\item Did you include the total amount of compute and the type of resources used (e.g., type of GPUs, internal cluster, or cloud provider)?
    \answerYes{yes, at section \cref{sec:experement_classification } and appendix C.}
     \item Do you justify how the proposed evaluation is sufficient and appropriate to the claims made? 
    \answerYes{yes, we report the significant test of all of the main experiments, Macnamer for classification comparison at \cref{sec:experement_classification }, and appendix C. Also, for the counter generation, we provide detailed validation process at \cref{sec:counter_gernration_experement }, and appendix D. }
     \item Do you discuss what is ``the cost`` of misclassification and fault (in)tolerance?
    \answerYes{yes, we report the significant test at appendix c, and indicates the limitation at \cref{sec:discussion}}
  
\end{enumerate}

\item Additionally, if you are using existing assets (e.g., code, data, models) or curating/releasing new assets, \textbf{without compromising anonymity}...
\begin{enumerate}
  \item If your work uses existing assets, did you cite the creators?
    \answerYes{yes, experiment set up at \cref{sec:dataset_Experemnt}}
  \item Did you mention the license of the assets?
    \answerYes{yes, models description at \cref{sec:experement_classification }, and appendix C. Also, dataset discussed at \cref{sec:dataset_Experemnt}}
  \item Did you include any new assets in the supplemental material or as a URL?
    \answerYes{yes, we include anonymous link to annotation guideline at appendix A }
  \item Did you discuss whether and how consent was obtained from people whose data you're using/curating?
    \answerYes{yes, we describe data labeling and collection at \cref{sec:dataset_Experemnt} and appendix A}
  \item Did you discuss whether the data you are using/curating contains personally identifiable information or offensive content?
    \answerYes{yes, as the domain of the study is counter hate speech analysis, thus we were cautious in including any official examples directly in the paper. instead, we use an optimal tool (Riveter) to analyze the discussion as we explained at ethical statement \cref{sec:ethical }}
\item If you are curating or releasing new datasets, did you discuss how you intend to make your datasets FAIR ?
\answerYes{yes,ethical statement \cref{sec:ethical }}
\item If you are curating or releasing new datasets, did you create a Datasheet for the Dataset ? 
\answerYes{yes, we prepare an electronic datasheet readme file at our private GitHub repository. We plan to share it after receiving the decision on the paper.}
\end{enumerate}

\item Additionally, if you used crowdsourcing or conducted research with human subjects, \textbf{without compromising anonymity}...
\begin{enumerate}
  \item Did you include the full text of instructions given to participants and screenshots?
    \answerYes{yes, detail annotation process at appendix A}
  \item Did you describe any potential participant risks, with mentions of Institutional Review Board (IRB) approvals?
    \answerYes{yes, detail annotation process at Appendix A, along with ethical statement \cref{sec:ethical }}
  \item Did you include the estimated hourly wage paid to participants and the total amount spent on participant compensation?
    \answerYes{yes, Annotation recruitment process at appendix A.}
   \item Did you discuss how data is stored, shared, and deidentified?
   \answerYes{yes, we discuss the data collection at appendix A. also we provide a description of the measures we used to protect the data at ethical statement \cref{sec:ethical }}
\end{enumerate}

\end{enumerate}


\appendix{}
\section{Preparation of the Baseline Datasets}
To prepare the DialogConan dataset, closed (multi-turn), we first unified topic definitions and mapped each target base topic into its main domain (racism, sexism, and religious bigotry) as follows:  
\\
$WOMEN + LGBT  \Rightarrow     SEXISM$ 
\\
$MIGRANTS +  POC    \Rightarrow   RACISM$  
\\
$MUSLIMS +  JEWS   \Rightarrow     RELIGIOUS BIGOTRY$
\\
\textcolor{black}{In general, the average number of tokens for each turn is around 19 tokens for the DiaolgConan (closed) and 27 tokens for the ContextCounter (open) dataset}. Moreover, to maintain the cost of the annotation, we selected 50\% of the dialogs of each topic in DialogConan. The dataset originally contained 3,059 unique dialogs spanning three topics, as shown in Table \ref{tab:appendexA_dilougs}.  This process results in our baseline dataset for the closed interaction dataset, which contains about 8,000 turns of conversations.

\begin{table}[h!]
\small
    \centering
    \begin{tabular}{c|c|c}
         Topic&  Unique dialogues& 50\% set of dialogues\\
         \hline
         Racisim&  1055& 527\\
         Sexism&  1031& 515\\
         Religious bigotry&  974& 487\\
    \end{tabular}
    \caption{Preparing DialogConan for annotation by selecting 50\% of the unique dialogues in each topic}
    \label{tab:appendexA_dilougs}
\end{table}

\section{Annotation Setup}
\begin{figure}[h!]
    \centering
    \includegraphics[width=0.95\linewidth]{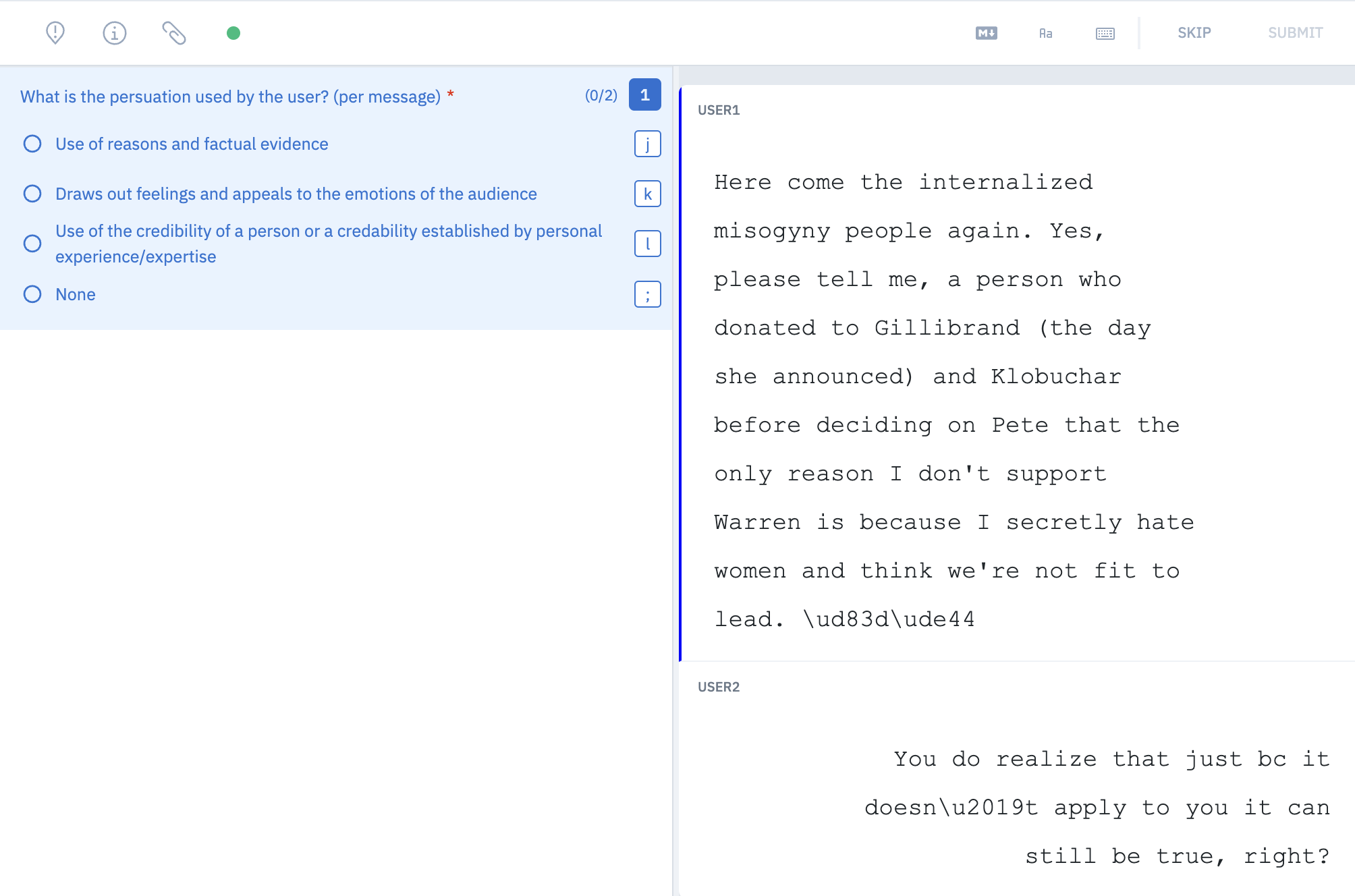}
    \caption{Annotation example for persuasion modes, where all the turns are shown to facilitate the labeling of each turn of the conversation}
    \label{fig:label_turn_labeling}
\end{figure}

\begin{figure}[h!]
    \centering
    \includegraphics[width=0.95\linewidth]{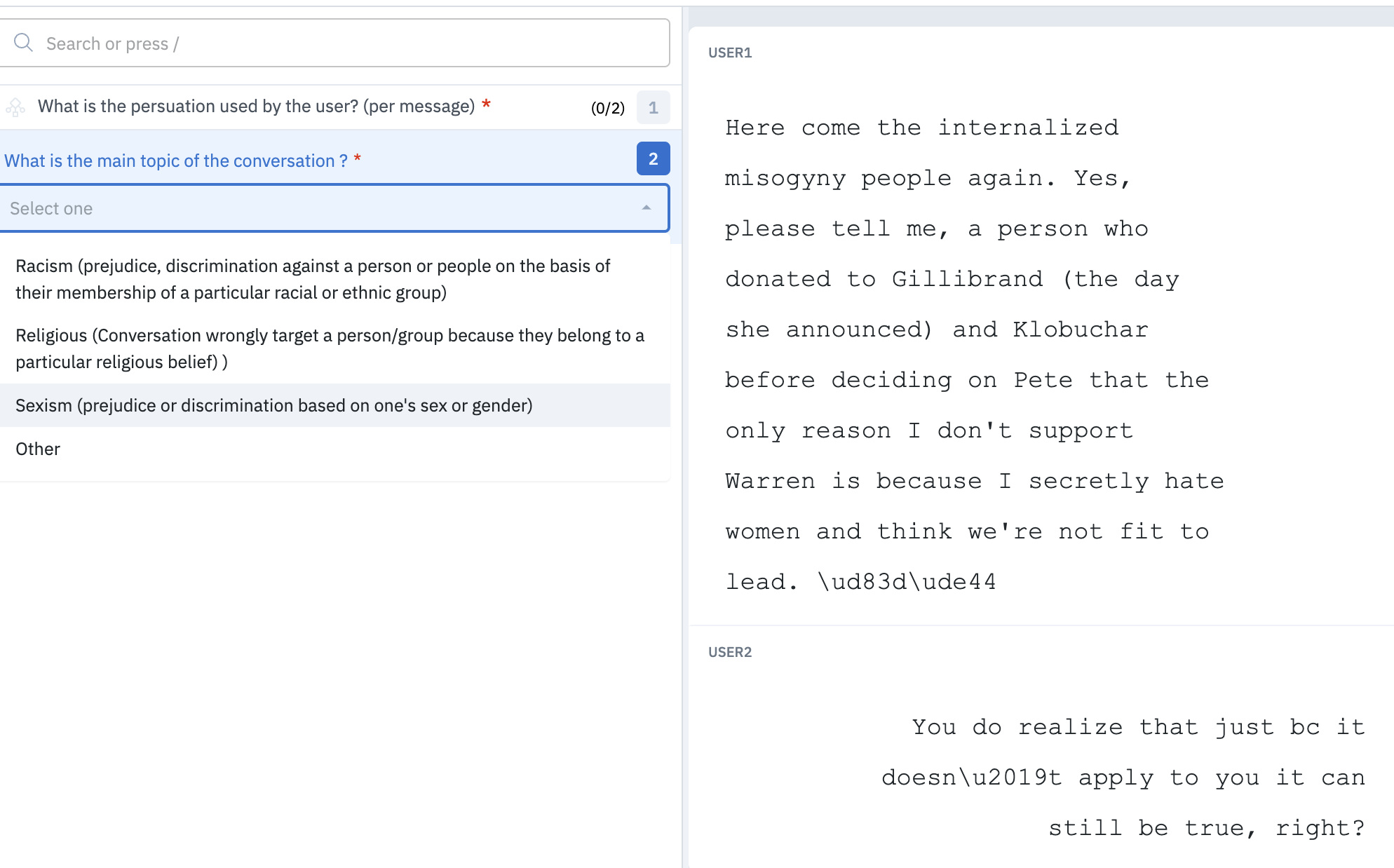}
    \caption{Topic annotation for Open (single-turn) dataset}
    \label{fig:topic_labeling}
\end{figure}

We use LabelBox as a platform to recruit and run annotation tasks. This platform enables conversation annotation, where the whole dialogue is shown to the annotator, with the possibility of labeling each conversation turn. We provide a thorough, detailed guideline for each interaction type (open/close) to the annotators, Figure~\ref{fig:label_turn_labeling}, and Figure~\ref{fig:topic_labeling}\footnote{\url{https://github.com/AbeerAldayel/counterspeech-persuasion-modes} }. A fair wage per hour has been provided to the team of two annotators as part of the Boost Tier payment setting of LabelBox. \footnote{\url{https://labelbox.com/product/annotate/}} 

\section{Classification of the Hate Speech and Counterspeech Model Hyperparameter} 

\textcolor{black}{Table \ref{tab:hyp} shows the major set of hyperparameters used for each model in the experiment. For the Uncased BERT-base, we used the pre-trained language models provided by the transformers Python library \footnote{\url{https://huggingface.co/transformers/v2.5.1/pretrained_models.html}}. Uncased BERT-base trained on lowered case English text with 12-layer, 768-hidden, 12-heads, 110M parameters. We also provided the details of dataset split for training, validation, and testing, in which we used a stratified split from scikit-learn and saved the split files to be used through the experiment to unify experiment settings and further ensure reproducibility as shown in Tables ~\ref{tab:split_multi} and ~\ref{tab:split_single}.}
\begin{table} [ht!]
\centering
\small
\begin{tabular}{l l l}
\hline
\textbf{\textbf{Model} } & \textbf{\textbf{Hyperparameter} } & \textbf{\textbf{Value} } \\
\hline
\multirow{4}{*}{\textbf{\textbf{SVM} }} & Regularization parameter(C)  & 3.0  \\

 & Kernal function  & rbf  \\

\multirow{5}{*}{\textbf{\textbf{Bilstm} }} & LSTM units (LSTM\_DIM)  & 64  \\
\hline
 & Epochs  & 10 \\
 & Spatial dropout  & 0.2  \\
 & Dropout rate  & 0.2  \\
 & Recurrent dropout  & 0.2  \\
 \hline
\multirow{6}{*}{\textbf{\textbf{BERT} }} & Epochs  & 5  \\
 & Train Batch size  & 16  \\
 & Eval Batch size  & 20  \\
 & Learning rate  & 5e-5  \\
 & Warm up steps  & 100  \\
 & Weight decay  & 0.01  \\
 \hline
\multirow{5}{*}{\textbf{\textbf{Llama-2-7B} }} & Batch size  & 1  \\
 & Learning rate  & 2e-4  \\
 & Epochs  & 20  \\
 & Max length  & 3  \\
 & Warm up steps  & 2  \\
 & lora alpha & 16 \\
 & lora dropout & 0.1 \\
 & lora rank & 64 \\
\hline
\end{tabular}
\caption{\textcolor{black}{Models hyperparameters }}
\label{tab:hyp}
\end{table}

\begin{table}
\small
\centering

\begin{tabular}{l l l l l}
\hline
\multicolumn{5}{l}{\textbf{\textbf{Closed (Multi-Turn)} }} \\
\hline
\textbf{\textbf{Topic} } & \textbf{Training}& \textbf{Validation }& \textbf{Test }& \textbf{Total}  \\
\hline
\textbf{\textbf{SEXISM} } & 1946  & 250  & 585  & 2781  \\

\textbf{RACISM } & 2069  & 266  & 622  & 2957  \\

\textbf{RELIGIOUS  } & 2011  & 258  & 605  & 2874  \\
\hline
\multicolumn{4}{l}{\textbf{\textbf{Count} }} & 8612  \\
\hline

\end{tabular}
\caption{Taring split for closed (multi-turn)}
\label{tab:split_multi}
\end{table}

\begin{table}
\small
\centering

\begin{tabular}{l l l l l}
\hline
\multicolumn{5}{l}{\textbf{\textbf{Open (Single-Turn)} }} \\
\hline
\textbf{\textbf{Topic} } & \textbf{Training }& \textbf{Validation }& \textbf{Test }& \textbf{Total}  \\
\hline
\textbf{\textbf{SEXISM} } & 384  & 49  & 116  & 549  \\

\textbf{RACISM } & 355  & 45  & 108  & 508  \\

\textbf{RELIGIOUS } & 195  & 25  & 59  & 279  \\
\hline
\multicolumn{4}{l}{\textbf{\textbf{Count} }} & 1336  \\
\hline

\end{tabular}
\caption{Taring split for open (single-turn)}
\label{tab:split_single}
\end{table}

For Llama-2 instruction tuning to classify a turn as hate or counter. In the model (text+per), we use the following instructions:
\quote{
"Below is an instruction that describes a task. Write a response that appropriately completes the request for the following persuasion setting. \\
setting: [persuasion mode] \\
Instruction: Classify the input text as Hate Speech or Counter Narrative \\
Input: [Text of turn]" \\
}
\\
\subsubsection{Significant Test of the Counter and Hate Predictions} \textcolor{black}{We use McNemar’s significant test ~\cite{McNEMAR1947-gd} to evaluate the significance of the baseline comparison on the model level. Table ~\ref{tab:sig_pred_macn} shows the results of comparing the significance of the model predictions where the model is trained on text only (txt) as well as on persuasion along with text (txt+pr). }

\begin{table}[!htbp]
\centering
\small
\renewcommand{\arraystretch}{1.2}
\setlength{\tabcolsep}{6pt}
\begin{tabular}{l l c}
\toprule
\textbf{Dataset} & \textbf{Model Comparison} & \textbf{p-value} \\
\midrule
\multirow{4}{*}{Closed (mT)} 
    & LLaMA-2 (txt vs. txt+pr) & \textbf{2.02e-11} **** \\
    & BERT (txt vs. txt+pr)    & 0.160 \\
    & BiLSTM (txt vs. txt+pr)  & \textbf{3.64e-06} **** \\
    & SVM (txt vs. txt+pr)     & 0.154 \\
\midrule
\multirow{4}{*}{Open (1T)} 
    & LLaMA-2 (txt vs. txt+pr) & \textbf{1.11e-16} **** \\
    & BERT (txt vs. txt+pr)    & 0.735 \\
    & BiLSTM (txt vs. txt+pr)  & \textbf{0.011} ** \\
    & SVM (txt vs. txt+pr)     & \textbf{0.00051} *** \\
\bottomrule
\end{tabular}
\caption{McNemar’s test results comparing model predictions when trained on text only (txt) versus text with persuasion mode features (txt+pr). Significance levels: ** $p<.01$, *** $p<.001$, **** $p<.0001$.}
\label{tab:sig_pred_macn}
\end{table}

\section{Generated Counterspeech Settings}
\subsubsection{Generating Counterspeech} 
We used task-specific prompts for Llama2-13B and GPT-3.5 Turbo to generate counterspeech for each message of hate.

\textbf{For GPT3,} we use the following prompt  \begin{displayquote}{"role": "system", "content": "You will be provided with a message (delimited with XML tags). Provide a counter-narrative to the arguments of each message. "},  
{"role": "user", "content": "$<message\> "+text+" </message>$"}, \end{displayquote}

\textbf{For Llama2,} we use the following prompt with input indicating a hate message 
\begin{displayquote} $[INST]$$ <<SYS>>$
provide a Counterargument to the input text below
$<</SYS>>$
${input}$
 $ [/INST]$
  \end{displayquote}

\begin{table*}[!htbp]
\centering
\small
\renewcommand{\arraystretch}{1.15}
\setlength{\tabcolsep}{4.5pt}

\begin{tabular}{l | l | l | l l l | l l l | l l l | l}
\toprule
\multirow{2}{*}{\textbf{Dataset}} & \multirow{2}{*}{\textbf{Model}} & \multirow{2}{*}{\textbf{Input}} 
& \multicolumn{3}{c|}{\textbf{Sexism}} 
& \multicolumn{3}{c|}{\textbf{Racism}} 
& \multicolumn{3}{c|}{\textbf{Religious}} 
& \textbf{Overall} \\
\cmidrule(lr){4-6} \cmidrule(lr){7-9} \cmidrule(lr){10-12}
& & & P & R & F1 & P & R & F1 & P & R & F1 & F1 \\
\midrule

\multicolumn{13}{l}{\textbf{Gold-Labeled (Closed + Open)}} \\
\midrule
& LLaMA-2-7B & txt       & 0.20 & 0.23 & 0.13 & 0.31 & 0.23 & 0.22 & 0.20 & 0.25 & 0.19 & 0.20 \\
& BERT       & txt       & 0.43 & 0.43 & 0.43 & \textbf{0.64} & \textbf{0.60} & \textbf{0.59} & \textbf{0.49} & 0.43 & \textbf{0.44} & \cellcolor{gray!15}\textbf{0.50} \\
& BiLSTM     & txt       & 0.35 & \textbf{0.83} & 0.49 & 0.19 & \textbf{0.85} & 0.32 & 0.36 & \textbf{0.88} & \textbf{0.51} & 0.33 \\
& SVM        & txt       & \textbf{0.47} & 0.42 & \textbf{0.43} & \textbf{0.66} & 0.54 & \textbf{0.57} & 0.42 & 0.39 & 0.38 & \cellcolor{gray!15}\textbf{0.50} \\

\midrule
\multicolumn{13}{l}{\textbf{Silver-Labeled (Closed - Multi-turn)}} \\
\midrule
& LLaMA-2-7B & txt       & \textbf{0.94} & \textbf{0.94} & \textbf{0.94} & 0.93 & 0.92 & 0.92 & 0.92 & 0.92 & 0.92 & \cellcolor{gray!15}\textbf{0.93} \\
&            & txt+pr    & 0.93 & 0.93 & 0.93 & \textbf{0.94} & \textbf{0.94} & \textbf{0.94} & 0.91 & 0.91 & 0.91 & \cellcolor{gray!15}\textbf{0.93} \\
& BERT       & txt       & \textbf{0.96} & \textbf{0.96} & \textbf{0.96} & 0.93 & 0.93 & 0.93 & \textbf{0.94} & \textbf{0.94} & \textbf{0.94} & \cellcolor{gray!15}\textbf{0.95} \\
&            & txt+pr    & 0.95 & 0.95 & 0.95 & \textbf{0.95} & \textbf{0.95} & \textbf{0.95} & 0.94 & 0.94 & 0.94 & \cellcolor{gray!15}\textbf{0.95} \\
& BiLSTM     & txt       & 0.88 & 0.88 & 0.88 & 0.85 & 0.85 & 0.85 & 0.88 & 0.87 & 0.87 & 0.87 \\
&            & txt+pr    & \textbf{0.89} & \textbf{0.89} & \textbf{0.89} & 0.85 & 0.85 & 0.85 & \textbf{0.88} & \textbf{0.88} & \textbf{0.88} & \cellcolor{gray!15}\textbf{0.88} \\
& SVM        & txt       & 0.90 & 0.90 & 0.90 & 0.88 & 0.88 & 0.88 & 0.90 & 0.90 & 0.90 & 0.89 \\
&            & txt+pr    & \textbf{0.90} & \textbf{0.90} & \textbf{0.90} & \textbf{0.88} & \textbf{0.88} & \textbf{0.88} & \textbf{0.91} & \textbf{0.91} & \textbf{0.91} & \cellcolor{gray!15}\textbf{0.90} \\

\bottomrule
\end{tabular}

\caption{Precision (P), Recall (R), and F1 scores for persuasion mode classification across datasets. Top: models trained on gold-labeled data (closed + open). Bottom: models trained on silver-labeled data (multi-turn only). }
\label{tab:combined_f1_scores_weaksuper}
\end{table*}

\subsubsection{Weak Supervision Training to Classify the Persuasion Modes in Counterspeech and Hate Speech } 
We trained four models on gold annotations from closed and open turns to classify the persuasion mode in the input turns (the text of hate speech). The classification output yielded three labels (emotion, reason, credibility), and we have reported the macro F1 score in Table ~\ref{tab:combined_f1_scores_weaksuper}. The goal is to identify and utilize the most effective model for assigning persuasion labels to the generated counter set. The outcomes presented in Table~\ref{tab:combined_f1_scores_weaksuper} show that BERT and SVM have the highest F1 score (50\%). On the contrary, Llama 2 has the lowest F1 score (20\%). This behavior of the instruction-tuning model can be shaped by the complexity of persuasion as a task (multi-classification and unpopular labeling). This has also been confirmed in a recent study in which they found that BERT provides an optimal result in comparison with other instruction-tuning models, such as Llama ~\citep{Thalken2023-ht}. By referring to the result in Table ~\ref{tab:combined_f1_scores_weaksuper}, on the topic level, BERT performs better on two topics out of the three (59\% F1 for the topic of racism and 44\% F1 for religious bigotry). Therefore, BERT has been used to populate the persuasion labels for the generated counter in closed and open datasets.


 \subsubsection{Validation of the Weak Supervision Model (Benchmark Method)} 
 \textcolor{black}{We used the weakly supervised model BERT to populate the rest of the DialogConan dataset (closed [multi-interaction]) with persuasion mode labels (silver labels). Then, we investigated the performance of the binary classification of hate speech and counterspeech using the same classification task in the section. However, we trained them on silver labels (weak labels).} Table \ref{tab:combined_f1_scores_weaksuper} shows the average F1 score of the three topics trained on the silver persuasion label to predict hate speech and counterspeech. For the instruction-tuning model (Llama 2), the F1 score remains the same at 93\% for text-only as well as text and persuasion scenarios. 
Similarly, the BERT model exhibits no change in the F1 score, maintaining a high accuracy of 95\%. The BiLSMT model sees a notable increase from 87\% to 88\% when incorporating persuasion modes. Furthermore, the baseline (SVM) model displays a similar pattern, with the F1 score improving from 89\% to 90\%. This consistent pattern across diverse models reinforces the weakly supervised model's efficacy in predicting the persuasion modes in the machine-generated counterspeech set.
\subsubsection{Validation of the Weak Supervision Model (Human Validation)}
\textcolor{black}{To validate the result of weakly supervised labels for the generated counter, we selected 50 generated counters from each topic and the generative model and validated the persuasion label. An independent annotator (one of the authors) performed the validation process through the two datasets to ensure consistency in the label moderation process. As shown in Table ~\ref{tab:error_rate}, in the closed dataset, the overall error rate is relatively small: about 19\% for GPT-3 and 12\% for Llama 2. For the open (single-turn) dataset, the overall error rate is about 22\% for GPT-3 and 16\% for Llama 2.  }    
 
\begin{table}[!ht]
\centering
\small
\renewcommand{\arraystretch}{1.2}
\setlength{\tabcolsep}{5pt}
\begin{tabular}{l | c c c | c}
\toprule
\textbf{Model} & \textbf{Racism} & \textbf{Sexism} & \shortstack{\textbf{Religious}} & \textbf{Overall} \\
\midrule
\multicolumn{5}{l}{\textbf{Closed ( Multi-turn)}} \\
GPT-3     & 22\% & 24\% & 12\% & \textbf{19\%} \\
LLaMA 2   &  2\% & 22\% & 12\% & \textbf{12\%} \\
\addlinespace
\midrule
\multicolumn{5}{l}{\textbf{Open ( Single-turn)}} \\
GPT-3     & 24\% & 22\% & 22\% & \textbf{22\%} \\
LLaMA 2   & 12\% & 20\% & 16\% & \textbf{16\%} \\
\bottomrule
\end{tabular}
\caption{Error rates (\%) in weak persuasion labels for machine-generated counterspeech across models. }
\label{tab:error_rate}
\end{table}

\end{document}